\documentclass[11pt,en,bibend=bibtex]{elegantpaper_cran}

\usepackage{graphicx}
\usepackage{booktabs}
\usepackage{amsmath,amssymb}
\usepackage{multirow}
\usepackage{colortbl}
\usepackage{xcolor}
\usepackage{pifont}
\usepackage{array}
\usepackage{caption}
\usepackage{algorithm}
\usepackage{algpseudocode}
\usepackage{placeins}
\usepackage{afterpage}
\usepackage{microtype}
\usepackage{hyperref}

\hypersetup{colorlinks=true,linkcolor=winered,citecolor=winered,urlcolor=winered}

\definecolor{bestrow}{RGB}{232,245,233}
\definecolor{oursrow}{RGB}{227,242,253}
\definecolor{gain}{RGB}{46,125,50}
\definecolor{loss}{RGB}{198,40,40}

\newcommand{\cmark}{\ding{51}}
\newcommand{\xmark}{\ding{55}}

\newcommand{\beginsupplement}{%
  \setcounter{section}{0}
  \renewcommand\thesection{S\arabic{section}}%
  \setcounter{table}{0}
  \renewcommand\thetable{S\arabic{table}}%
  \setcounter{figure}{0}
  \renewcommand\thefigure{S\arabic{figure}}%
  \setcounter{algorithm}{0}
  \renewcommand\thealgorithm{S\arabic{algorithm}}%
}

\title{Cross-Resolution Attention Network for\\
High-Resolution PM$_{2.5}$ Prediction}

\author{%
  Ammar Kheder\textsuperscript{2,3}\thanks{Corresponding: \texttt{ammar.kheder@lut.fi}}
  \quad
  Helmi Toropainen\textsuperscript{1,3}
  \quad
  Wenqing Peng\textsuperscript{1,3}
  \quad
  Samuel Ant\~{a}o\textsuperscript{4}
  \quad
  Zhi-Song Liu\textsuperscript{2,3,**}
  \quad
  Michael Boy\textsuperscript{1,2,3}\\[4pt]
  \small
  \textsuperscript{1}Institute for Atmospheric and Earth System Research, University of Helsinki, P.O. Box 64, Helsinki 00014, Finland\\
  \textsuperscript{2}Department of Computational Engineering, LUT University, Finland\\
  \textsuperscript{3}Atmospheric Modelling Centre Lahti (AMC-Lahti), Finland\\
  \textsuperscript{4}Advanced Micro Devices (AMD), Munich, Germany
}

\date{}


\begin{document}
\maketitle
{\let\thefootnote\relax\footnotetext{\textsuperscript{**}Corresponding: \texttt{zhisong.liu@lut.fi}}}

\begin{abstract}
Vision Transformers have achieved remarkable success in spatio-temporal prediction,
but their scalability remains limited for ultra-high-resolution, continent-scale domains
required in real-world environmental monitoring. A single European air-quality map at
1\,km resolution comprises 29~million pixels, far beyond the limits of naive self-attention.
We introduce \textbf{CRAN-PM}, a dual-branch Vision Transformer that leverages
cross-resolution attention to efficiently fuse global meteorological data (25\,km) with
local high-resolution PM$_{2.5}$ at the current time (1\,km). Instead of including
physically driven factors like temperature and topography as input, we further introduce
elevation-aware self-attention and wind-guided cross-attention to force the network to
learn physically consistent feature representations for PM$_{2.5}$ forecasting.
CRAN-PM is fully trainable and memory-efficient, generating the complete 29-million-pixel
European map in 1.8\,seconds on a single GPU. Evaluated on daily PM$_{2.5}$ forecasting
throughout Europe in 2022 (362~days, 2{,}971 European Environment Agency (EEA) stations),
it reduces RMSE by 4.7\% at T+1 and 10.7\% at T+3 compared to the best single-scale
baseline, while reducing bias in complex terrain by 36\%.

\end{abstract}

\noindent\textbf{Code:} \url{https://github.com/AmmarKheder/cran_pm}

\section{Introduction}
\label{sec:intro}

Vision Transformers~\cite{dosovitskiy2021image} have transformed image and video
processing for spatio-temporal analysis, with models such as Swin
Transformer~\cite{liu2021swin}, VPTR~\cite{vptr}, ViViT~\cite{vivit},
Earthformer~\cite{gao2022earthformer}, and ClimaX~\cite{nguyen2023climax} achieving
state-of-the-art results across video prediction, weather forecasting, and remote sensing.
Yet all of these methods share a fundamental limitation: they operate at a single, fixed
spatial resolution. When the target domain is large and the required resolution is fine,
the token count explodes and the model becomes intractable. This is not merely an
engineering inconvenience; it is a structural barrier preventing Vision Transformers from
addressing real-world problems requiring both ultra-high resolution and long-range spatial
context.

\begin{figure}[!ht]
  \centering
  \includegraphics[width=\textwidth]{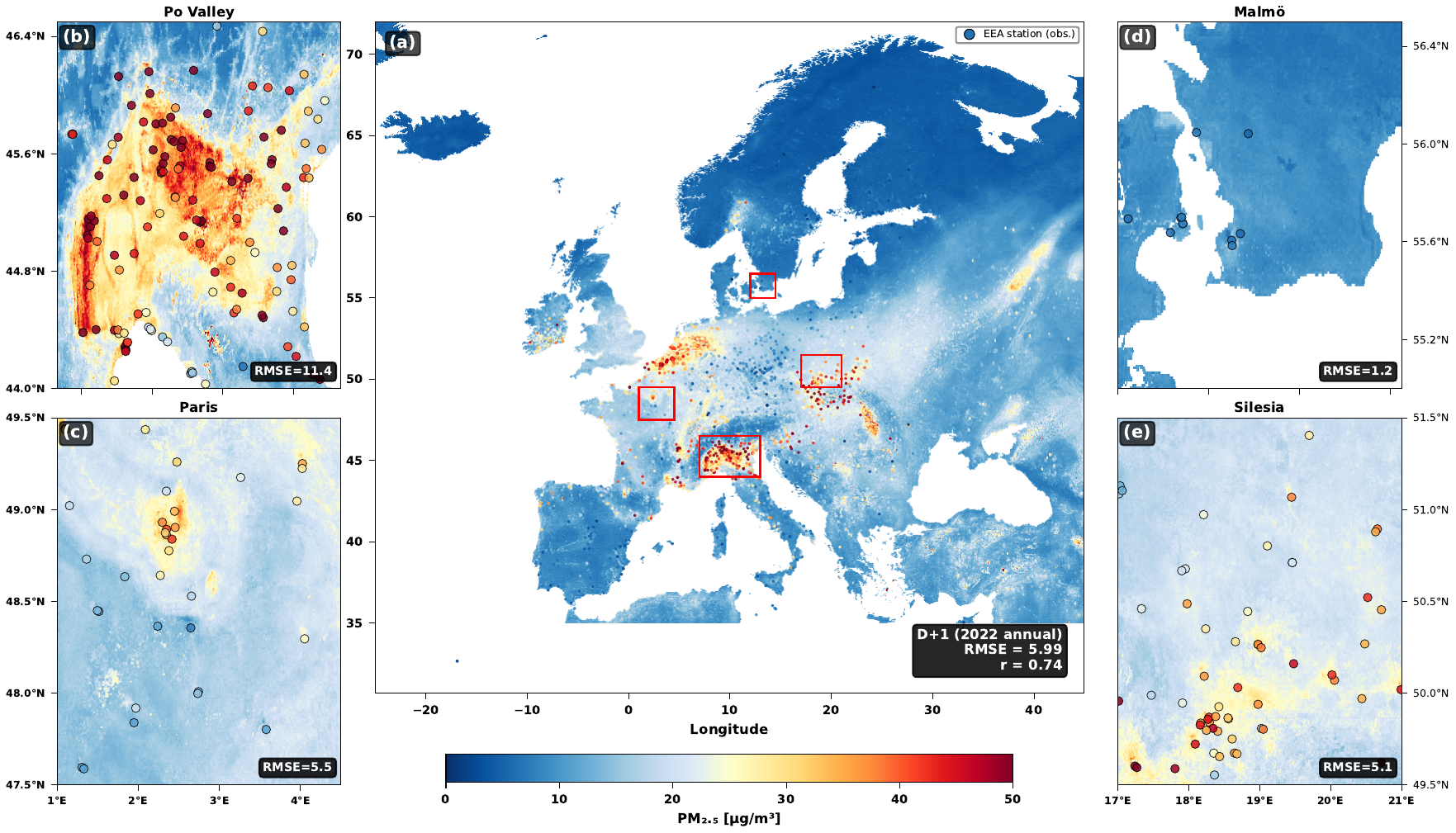}
  \caption{\textbf{CRAN-PM predicts daily PM$_{2.5}$ at 1\,km resolution across Europe.}
  (a)~Full European prediction for January~25, 2022 (T+1 horizon).
  Red rectangles indicate zoom regions.
  (b--e)~Regional details for Po Valley, Paris, Malm\"o, and Silesia.
  Colored circles represent independent EEA ground station measurements.
  Annual T+1 performance: RMSE\,=\,5.99\,\textmu g/m$^3$, $r$\,=\,0.74.}
  \label{fig:teaser}
\end{figure}

Our goal is to design a PM$_{2.5}$ forecasting model that takes the current observation
to predict future evolution at fine resolution (1\,km$\times$1\,km). However, a European
map at this resolution is approximately $4000\!\times\!8000$ pixels, yielding
${\sim}$115{,}000 tokens when tokenized, far exceeding memory limits for standard
self-attention. Independent tiling discards the large-scale meteorological context
(advection, boundary-layer stability) driving local pollution. Existing methods either
maintain global consistency at coarse resolutions
($\geq$10--40\,km)~\cite{cams,bi2023pangu,lam2023graphcast,bodnar2024aurora,topoflow},
or provide retrospective 1\,km estimates without forecasting
capabilities~\cite{wei2023ghap}. The central question is: \textit{how can a Vision
Transformer handle ultra-high-resolution forecasting over continental domains without
losing global physical context?}

We introduce \textbf{CRAN-PM} (Fig.~\ref{fig:architecture}), a dual-branch Vision
Transformer designed for scalable, high-resolution air-quality forecasting. A
\textbf{global branch} encodes large-scale meteorology over the full domain at 25\,km
resolution, while a \textbf{local branch} processes overlapping 1\,km tiles on the
current high-resolution PM$_{2.5}$ data. The branches communicate via \textbf{cross-resolution
attention}, enabling the local branch to query the global branch for long-range context.
This reduces per-tile memory to under 2\,GB while preserving physically meaningful
interactions. We further propose elevation-aware self-attention bias and a wind-guided
cross-attention bias as soft physical constraints.

Our contributions are:
\begin{itemize}
  \item Existing methods either upsample coarse meteorology or downsample fine-resolution
  PM$_{2.5}$ to enforce uniform inputs, discarding spatial detail or introducing
  interpolation artefacts. We instead propose a dual-branch Vision Transformer that
  processes global meteorological fields and local PM$_{2.5}$ patches at their native
  resolutions, bridging them via cross-attention. This lets the model learn which
  atmospheric drivers most influence each local region without any resolution compromise.
  \item We incorporate elevation and wind-field priors as soft additive biases in the
  attention, grounding the model in known physical transport mechanisms. These priors
  require no additional parameters and improve accuracy at complex-terrain stations
  where data-driven baselines systematically fail.
  \item To our knowledge, this is the first deep-learning model to produce Europe-wide
  PM$_{2.5}$ forecasts at 1\,km spatial resolution. Evaluated on 2{,}971 EEA stations
  across 362 days, CRAN-PM achieves RMSE\,=\,6.85\,\textmu g/m$^3$ at T+1, outperforming
  all baselines by 4.7--10.7\% and reducing bias at complex-terrain stations by 36\%.
\end{itemize}

\section{Related Work}
\label{sec:related}

\noindent\textbf{Vision Transformers for high-resolution and multi-scale prediction.}
ViT~\cite{dosovitskiy2021image} demonstrated that pure self-attention over image patches
matches convolutional networks at scale, but its $\mathcal{O}(n^2)$ complexity limits
large images. Swin Transformer~\cite{liu2021swin} introduced shifted-window attention
with linear complexity, enabling higher-resolution inputs but restricting receptive fields
locally. PVT~\cite{wang2021pvt} and Twins~\cite{chu2021twins} reduce quadratic cost via
spatial-reduction attention while preserving multi-scale features. Video
Swin~\cite{liu2022video} and TimeSformer~\cite{bertasius2021spacetime} extend attention
to spatio-temporal data but remain limited to moderate resolutions.
CrossViT~\cite{chen2021crossvit} exchanges information between two scales via class
tokens, yet both branches operate on the same image domain. HIPT~\cite{chen2022hipt}
and TransMIL~\cite{shao2021transmil} process patches independently without cross-scale
interaction. CRAN-PM overcomes this via cross-resolution attention, allowing local
high-resolution tokens to query a separately encoded global representation.

\noindent\textbf{Data-driven weather and air quality prediction.}
AI weather models such as Pangu-Weather~\cite{bi2023pangu},
GraphCast~\cite{lam2023graphcast}, FourCastNet~\cite{pathak2022fourcastnet},
ClimaX~\cite{nguyen2023climax}, and Aurora~\cite{bodnar2024aurora} achieve strong
forecasting skill at 0.1--0.25$^\circ$ resolution. FuXi~\cite{chen2023fuxi} and
GenCast~\cite{price2024gencast} extend horizons but remain $\geq$0.25$^\circ$.
Satellite products such as GHAP~\cite{wei2023ghap} provide retrospective 1\,km
PM$_{2.5}$ estimates, and CAMS~\cite{cams} operates near 0.4$^\circ$. Station-based
models using gradient-boosted trees~\cite{chen2019xgboost} or
GNNs~\cite{wang2020pm25gnn} lack spatial continuity.
CRAN-PM bridges the gap by enabling 1\,km continental prediction through explicit
cross-resolution attention.

\noindent\textbf{Physics-guided deep learning.}
Physical knowledge has been incorporated via PDE-constrained
losses~\cite{raissi2019physics,spinode,nne}, latent physics-residual
decompositions~\cite{guen2020phydnet}, hybrid advection-generation
models~\cite{zhang2023nowcastnet}, and Fourier Neural
Operators~\cite{li2021fno,li2024pino}. Building on single-scale physics-guided
models TopoFlow~\cite{topoflow} and AQ-Net~\cite{kheder2025aqnet}, we encode physics
as architectural inductive biases: elevation-aware attention and wind-guided
cross-attention inject topographic and advective structure directly within attention.

\section{Method}
\label{sec:method}

\subsection{Cross-Resolution Attention for Ultra-High-Resolution ViTs}
\label{sec:divide}

Forecasting PM$_{2.5}$ at European scale and 1\,km resolution requires processing
${\sim}$29 million pixels per map. Tokenizing with $16\!\times\!16$ patches yields
${\sim}$115{,}000 tokens, making naive self-attention intractable. Windowed attention
(e.g., Swin~\cite{liu2021swin}) is feasible but confines tokens locally, discarding
the large-scale meteorological context essential for day-to-day pollution dynamics.

We adopt a divide-and-conquer strategy: coarse meteorological data (25\,km) and fine
PM$_{2.5}$ (1\,km) are split into global tokens and local tiles. Our
\textbf{cross-resolution attention} lets local tokens query global tokens for
large-scale context, preserving long-range dependencies while reducing memory to
under 2\,GB per tile. Inference over the full 29-million-pixel pan-European map
completes in 1.8\,seconds on a single GPU.

\begin{figure}[!ht]
  \centering
  \includegraphics[width=\textwidth]{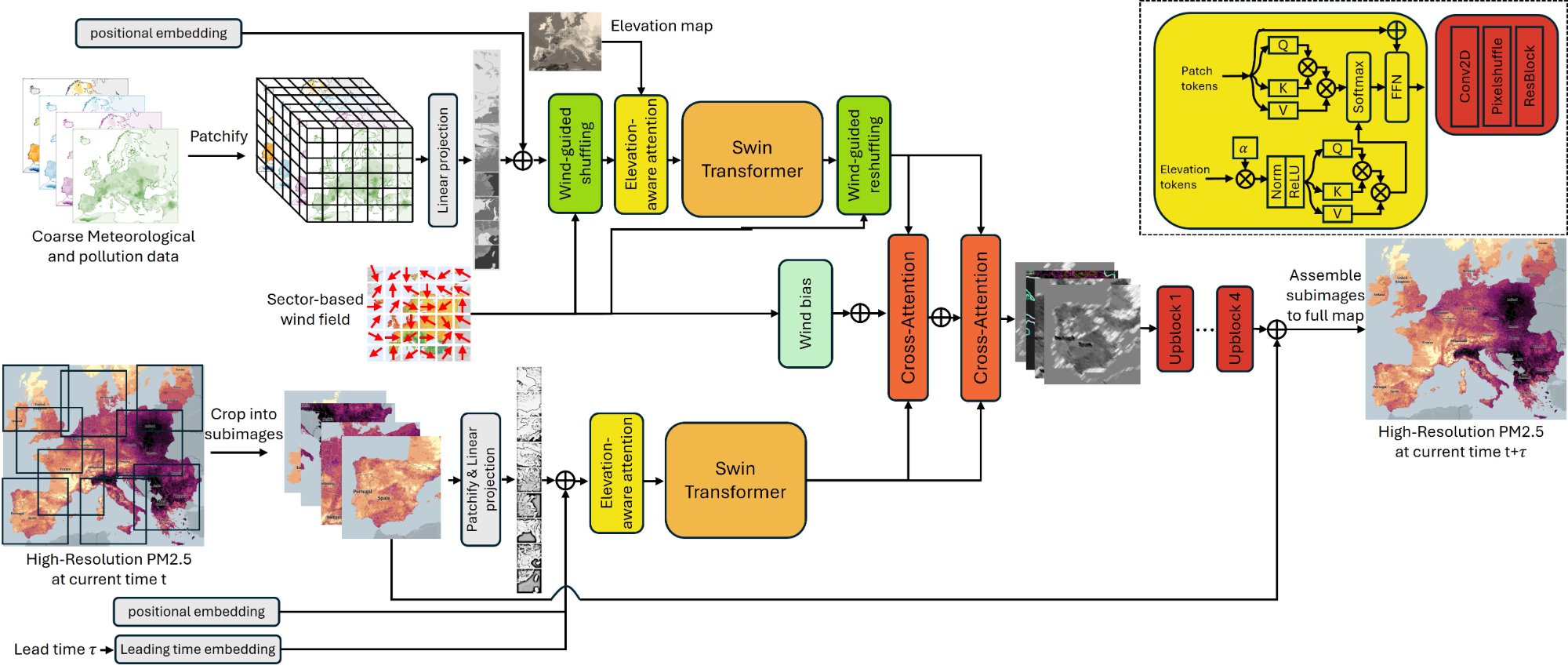}
  \caption{\textbf{Architecture of CRAN-PM.} A global branch (top) encodes coarse
  meteorological fields with wind-guided token reordering and elevation-aware attention;
  a local branch (bottom) encodes high-resolution PM$_{2.5}$ subimages. Two wind-biased
  cross-attention layers fuse the branches (fine queries coarse). A PixelShuffle-based
  Upblock (red inset) reconstructs the residual. The yellow inset details elevation-aware
  attention.}
  \label{fig:architecture}
\end{figure}

Fig.~\ref{fig:architecture} shows CRAN-PM's overall architecture: (1) the
\textbf{global branch} takes coarse meteorological data across Europe (25\,km,
168$\times$280 patches, 735 tokens); (2) the \textbf{local branch} processes
current-day fine-resolution PM$_{2.5}$ (1\,km) into 126 overlapping
512$\times$512 tiles; (3) the \textbf{cross-attention module} learns long-range
correlations between local PM$_{2.5}$ and global meteorology; (4) upsampling
blocks reconstruct the residual, added to today's PM$_{2.5}$ observation.

\subsection{Problem Formulation}
\label{sec:formulation}

Let $\mathbf{x}^{\text{g}}_t \in \mathbb{R}^{70 \times 168 \times 280}$ denote the
coarse input at day $t$, comprising ERA5 reanalysis (60 channels at days $t$ and
$t{-}1$) concatenated with CAMS composition forecasts (10 channels) at 0.25$^\circ$.
The fine input $\mathbf{x}^{\ell}_t \in \mathbb{R}^{5 \times 512 \times 512}$ is a
local tile of fine-resolution PM$_{2.5}$ at days $t$ and $t{-}1$, plus elevation,
latitude, and longitude at 0.01$^\circ$. For lead time $\tau \in \{1,2,3\}$ days:
\begin{equation}
\hat{y}_{t+\tau} = x^{\ell}_t + f_\theta\!\left(\mathbf{x}^{\text{g}}_t,
\mathbf{x}^{\ell}_t, \tau\right),
\label{eq:delta}
\end{equation}
where $f_\theta$ outputs a residual map $\Delta$ modelling daily changes. This delta
formulation initialises the network near persistence, simplifying
learning~\cite{bi2023pangu,lam2023graphcast}.

\subsection{Multi-Scale Encoder}
\label{sec:encoder}

\paragraph{Global branch.}
The coarse input is partitioned into $8\!\times\!8$ patches yielding $N_g = 735$
tokens projected to $d_g = 768$. Sinusoidal 2D positional embeddings are added,
followed by one elevation-aware attention block (Sec.~\ref{sec:eaa}) and one Swin
Transformer.

\paragraph{Local branch.}
Each $512\!\times\!512$ tile is tokenized with $16\!\times\!16$ patches, producing
$N_\ell = 1{,}024$ tokens at $d_\ell = 512$. A learnable lead-time embedding
$\mathbf{E}_{\text{lead}}(\tau) \in \mathbb{R}^{512}$ is added, followed by one
elevation-aware attention block ($E_0 = 500$\,m) and one Swin Transformer.

\paragraph{Wind-guided shuffling.}
Before the global transformer, each $7\!\times\!7$ patch group is reordered according
to the local wind field. The ERA5 mean wind vector is quantized into one of 16
directional sectors ($22.5^\circ$ each), and patches are scanned upwind to downwind,
aligning sequential processing with the physical advection path~\cite{topoflow}.

\subsection{Elevation-Aware Attention}
\label{sec:eaa}

For token pair $(i,j)$ with mean elevations $E_i$ and $E_j$:
\begin{equation}
B^{\text{elev}}_{ij} = -\alpha \cdot \text{ReLU}\!\Big(\frac{E_j - E_i}{E_0}\Big),
\quad B^{\text{elev}}_{ij} \in [-10, 0],
\end{equation}
where $\alpha$ is learnable and $E_0 = 1{,}000$\,m (global) or 500\,m (local).
This asymmetric bias penalizes attention to higher-elevation sources, consistent
with katabatic flows~\cite{whiteman2000mountain}:
\begin{equation}
\text{Attn}_{\text{elev}} = \text{softmax}\!\Big(\frac{\mathbf{Q}\mathbf{K}^\top}
{\sqrt{d_h}} + \mathbf{B}^{\text{elev}} + \mathbf{B}^{\text{rel}}\Big)\mathbf{V}.
\end{equation}

\subsection{Wind-Guided Cross-Attention}
\label{sec:cross_attn}

Local tokens query global tokens for large-scale context. Global tokens are projected
to match local dimension: $\tilde{\mathbf{Z}}^{\text{g}} = \text{Linear}(\mathbf{Z}^{\text{g}})
\in \mathbb{R}^{735 \times 512}$. A wind-guided bias encodes upwind alignment:
\begin{equation}
\gamma_{ij} = \frac{(\mathbf{p}_i - \mathbf{p}_j)}{\|\mathbf{p}_i - \mathbf{p}_j\|}
\cdot \frac{(u_{10,i}, v_{10,i})}{\|(u_{10,i}, v_{10,i})\|}, \quad
B^{\text{wind}}_{ij} = \beta \cdot \gamma_{ij},
\end{equation}
\begin{equation}
\text{CrossAttn} = \text{softmax}\Big(\frac{\mathbf{Q}\mathbf{K}^\top}{\sqrt{d_h}}
+ \mathbf{B}^{\text{wind}}\Big)\mathbf{V},
\end{equation}
where $\beta$ is learnable per head. Two successive cross-attention layers with
residual connections and FFNs progressively integrate coarse-scale context.

\subsection{Decoder}
\label{sec:decoder}

Fused local tokens are reshaped to $512\!\times\!32\!\times\!32$. Four upsampling
blocks (Conv + PixelShuffle~\cite{shi2016pixelshuffle} + residual) progressively
restore $512\!\times\!512$ resolution. A final zero-initialized $1\!\times\!1$
convolution outputs residual $\Delta$, added to $x^{\ell}_t$.

\subsection{Training Objective}
\label{sec:loss}

\begin{equation}
\mathcal{L} = \mathcal{L}_{\text{pixel}} + \lambda_{\text{FFL}}\,\mathcal{L}_{\text{FFL}}
+ \lambda_{\text{station}}\,\mathcal{L}_{\text{station}},
\end{equation}
where $\mathcal{L}_{\text{pixel}}$ is MSE over land pixels, $\mathcal{L}_{\text{FFL}}$
is the focal frequency loss~\cite{jiang2021focal} emphasizing high-frequency errors, and
$\mathcal{L}_{\text{station}}$ anchors predictions to $N_s = 2{,}971$ EEA stations.
We set $\lambda_{\text{FFL}} = \lambda_{\text{station}} = 0.1$.

\begin{figure}[!ht]
  \centering
  \includegraphics[width=\textwidth]{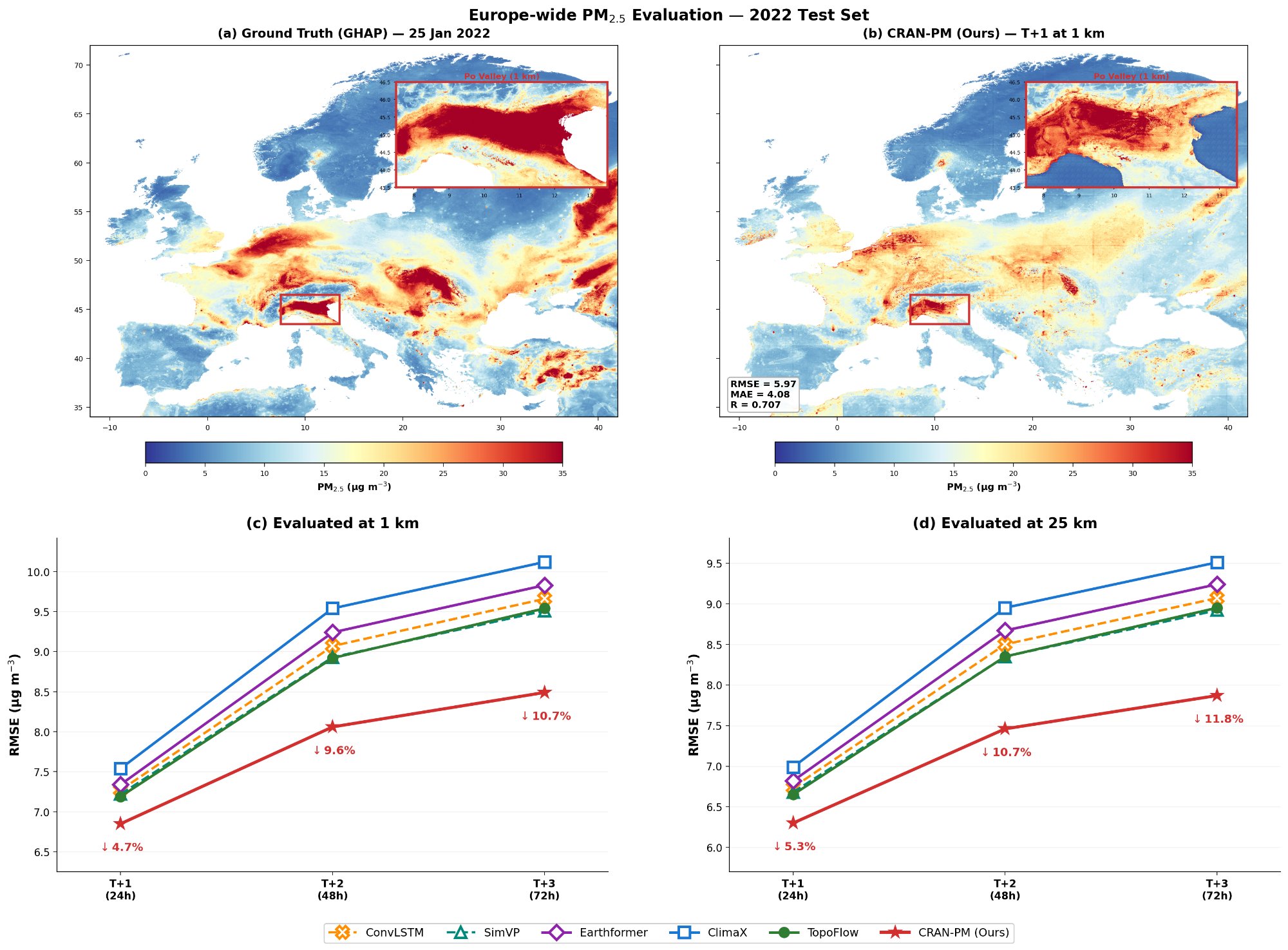}
  \caption{\textbf{Europe-wide PM$_{2.5}$ evaluation (2022).}
  (a)~GT (GHAP, 1\,km), Jan.~25, 2022. (b)~CRAN-PM T+1; inset: Po Valley.
  (c,d)~RMSE across T+1--T+3 at 1\,km and 25\,km.
  CRAN-PM (red stars) consistently outperforms all baselines.}
  \label{fig:europe_maps}
\end{figure}

\section{Experiments}
\label{sec:experiments}

\subsection{Data}
\label{sec:data}

Our domain covers Europe (33$^\circ$N--72$^\circ$N, 25$^\circ$W--45$^\circ$E) at
1\,km resolution (4{,}192$\times$6{,}992 pixels, 126 overlapping $512\!\times\!512$
tiles). Coarse inputs ($C_g = 70$ channels, $\sim$25\,km) combine \textbf{ERA5
reanalysis} (60 channels at days $t$ and $t{-}1$) and \textbf{CAMS
analysis}~\cite{cams} (10 channels at days $t$ and $t{-}1$). Fine inputs
($C_\ell = 5$ channels, $\sim$1\,km) comprise \textbf{GHAP}~\cite{wei2023ghap}
satellite-derived PM$_{2.5}$ at $t$ and $t{-}1$, plus SRTM elevation, latitude,
longitude. We train on 2017--2021 and test on 2022 (362 days, strictly temporal
split, zero leakage).

\subsection{Implementation Details}
\label{sec:implementation}

Global branch: $d_g = 768$, 8 blocks (1 elevation-aware + 7 Swin), 12 heads,
patch $8\!\times\!8$. Local branch: $d_\ell = 512$, 6 blocks (1 + 5), 8 heads,
patch $16\!\times\!16$. Cross-attention bridge: 2 layers, 8 heads, head dim 64.
Total: 96\,M parameters. Training: 64 AMD MI250X GPUs (LUMI supercomputer), 30 epochs,
AdamW ($\text{lr}=5\!\times\!10^{-5}$), cosine schedule, bfloat16 mixed precision,
$\approx$860 GPU-hours. Inference: 1.8\,s/map on one MI250X GPU.

\subsection{Baselines}

We compare against CAMS$^\dagger$~\cite{cams}, ConvLSTM~\cite{shi2015convolutional},
SimVP~\cite{gao2022simvp}, Earthformer~\cite{gao2022earthformer},
ClimaX~\cite{nguyen2023climax}, and TopoFlow~\cite{topoflow}. All learned baselines
operate at 25\,km and are bilinearly interpolated to 1\,km for high-resolution
evaluation; for 25\,km evaluation, CRAN-PM is spatially averaged.

\subsection{Main Results}
\label{sec:results}

\begin{table}[!ht]
\centering
\setlength{\tabcolsep}{4pt}
\caption{\textbf{Quantitative comparison on European PM$_{2.5}$ prediction.}
\textit{Top}: evaluated at 1\,km (coarser methods bilinearly interpolated).
\textit{Bottom}: evaluated at 25\,km (CRAN-PM spatially averaged).
$\dagger$: zero-shot. \textbf{Bold}: best. \underline{Underline}: second best.}
\label{tab:main}
\resizebox{\textwidth}{!}{%
\begin{tabular}{@{}lcc rrrr rrrr rrrr@{}}
\toprule
& & &
\multicolumn{4}{c}{\textbf{T+1 (24\,h)}} &
\multicolumn{4}{c}{\textbf{T+2 (48\,h)}} &
\multicolumn{4}{c}{\textbf{T+3 (72\,h)}} \\
\cmidrule(lr){4-7}\cmidrule(lr){8-11}\cmidrule(lr){12-15}
\textbf{Method} & \textbf{Res.} & \textbf{\#Param} &
RMSE$\downarrow$ & MAE$\downarrow$ & SSIM$\uparrow$ & $\Delta$\,(\%) &
RMSE$\downarrow$ & MAE$\downarrow$ & SSIM$\uparrow$ & $\Delta$\,(\%) &
RMSE$\downarrow$ & MAE$\downarrow$ & SSIM$\uparrow$ & $\Delta$\,(\%) \\
\midrule
\multicolumn{15}{@{}l}{\textit{Evaluated at 1\,km resolution}} \\
\midrule
CAMS$^\dagger$~\cite{cams}            & 1\,km  & --    & 12.35 & 7.34 & 0.38 & -- & 12.41 & 7.45 & 0.36 & -- & 12.45 & 7.51 & 0.34 & -- \\
ConvLSTM~\cite{shi2015convolutional}  & 25\,km & 8.2M  & 7.27  & 3.24 & 0.48 & -- & 9.07  & 4.08 & 0.45 & -- & 9.66  & 4.44 & 0.42 & -- \\
SimVP~\cite{gao2022simvp}             & 25\,km & 4.9M  & 7.22  & 3.27 & 0.51 & -- & 8.93  & 4.05 & 0.48 & -- & \underline{9.51}  & \underline{4.40} & 0.45 & -- \\
Earthformer~\cite{gao2022earthformer} & 25\,km & 61M   & 7.34  & 3.31 & 0.49 & -- & 9.24  & 4.14 & 0.46 & -- & 9.83  & 4.50 & 0.44 & -- \\
ClimaX~\cite{nguyen2023climax}        & 25\,km & 57M   & 7.54  & 3.22 & 0.47 & -- & 9.54  & 4.15 & 0.44 & -- & 10.12 & 4.52 & 0.41 & -- \\
TopoFlow~\cite{topoflow}              & 25\,km & 61M   & \underline{7.19} & \underline{3.31} & \underline{0.53} & -- & \underline{8.92} & \underline{4.10} & \underline{0.50} & -- & 9.54 & 4.47 & \underline{0.47} & -- \\
\rowcolor{oursrow}
\textbf{CRAN-PM (ours)}               & \textbf{1\,km} & \textbf{96M} &
\textbf{6.85} & \textbf{3.14} & \textbf{0.78} & $\downarrow$\textbf{4.7} &
\textbf{8.06} & \textbf{3.79} & \textbf{0.74} & $\downarrow$\textbf{9.6} &
\textbf{8.49} & \textbf{4.07} & \textbf{0.71} & $\downarrow$\textbf{10.7} \\
\midrule
\multicolumn{15}{@{}l}{\textit{Evaluated at 25\,km resolution}} \\
\midrule
CAMS$^\dagger$~\cite{cams}            & 25\,km & --    & 11.96 & 7.13 & 0.58 & -- & 12.02 & 7.23 & 0.55 & -- & 12.05 & 7.28 & 0.53 & -- \\
ConvLSTM~\cite{shi2015convolutional}  & 25\,km & 8.2M  & 6.74  & 2.80 & 0.74 & -- & 8.50  & 3.60 & 0.70 & -- & 9.07  & 3.95 & 0.67 & -- \\
SimVP~\cite{gao2022simvp}             & 25\,km & 4.9M  & 6.68  & 2.83 & 0.77 & -- & 8.35  & 3.57 & 0.73 & -- & \underline{8.92} & \underline{3.90} & 0.69 & -- \\
Earthformer~\cite{gao2022earthformer} & 25\,km & 61M   & 6.82  & 2.91 & 0.75 & -- & 8.67  & 3.69 & 0.71 & -- & 9.24  & 4.03 & 0.68 & -- \\
ClimaX~\cite{nguyen2023climax}        & 25\,km & 57M   & 6.99  & 2.72 & 0.73 & -- & 8.95  & 3.62 & 0.69 & -- & 9.51  & 3.97 & 0.65 & -- \\
TopoFlow~\cite{topoflow}              & 25\,km & 61M   & \underline{6.65} & \underline{2.85} & \underline{0.79} & -- & \underline{8.35} & \underline{3.61} & \underline{0.75} & -- & 8.95 & 3.97 & \underline{0.72} & -- \\
\rowcolor{oursrow}
\textbf{CRAN-PM (ours)}               & \textbf{1\,km} & \textbf{96M} &
\textbf{6.30} & \textbf{2.85} & \textbf{0.84} & $\downarrow$\textbf{5.3} &
\textbf{7.46} & \textbf{3.46} & \textbf{0.80} & $\downarrow$\textbf{10.7} &
\textbf{7.87} & \textbf{3.72} & \textbf{0.76} & $\downarrow$\textbf{11.8} \\
\bottomrule
\end{tabular}}
\end{table}

Table~\ref{tab:main} shows that at 1\,km, CRAN-PM achieves RMSE\,=\,6.85 at T+1
($-$4.7\% vs.\ TopoFlow 7.19). The improvement widens at longer horizons: $-$9.6\%
at T+2 and $-$10.7\% at T+3. SSIM of 0.78 vs.\ 0.53 reflects fine-scale spatial
structure that interpolated 25\,km models cannot capture. At 25\,km, CRAN-PM remains
the best method ($-$5.3\% RMSE at T+1), confirming improvements stem from better
predictions rather than resolution alone.

Table~\ref{tab:stations} reports validation against EEA ground stations.
\begin{table}[!ht]
\centering
\setlength{\tabcolsep}{4pt}
\caption{\textbf{Validation against EEA ground monitoring stations (2022, T+1).}
Stations grouped by terrain complexity ($\sigma_z$: elevation s.d.\ within 25\,km radius).
\textbf{Bold}: best. $\dagger$: physical model.}
\label{tab:stations}
\resizebox{\textwidth}{!}{%
\begin{tabular}{@{} l rrrr c rrrr c rrrr @{}}
\toprule
&
\multicolumn{4}{c}{\textbf{All Stations} ($N$\,=\,2971)} & &
\multicolumn{4}{c}{\textbf{Flat} ($\sigma_z <$ 50\,m, $N$\,=\,2096)} & &
\multicolumn{4}{c}{\textbf{Complex} ($\sigma_z \geq$ 50\,m, $N$\,=\,875)} \\
\cmidrule(lr){2-5}\cmidrule(lr){7-10}\cmidrule(lr){12-15}
\textbf{Method} &
RMSE$\downarrow$ & MAE$\downarrow$ & SSIM$\uparrow$ & Bias & &
RMSE$\downarrow$ & MAE$\downarrow$ & SSIM$\uparrow$ & Bias & &
RMSE$\downarrow$ & MAE$\downarrow$ & SSIM$\uparrow$ & Bias \\
\midrule
CAMS$^\dagger$                        & 13.08 & 6.16 & 0.52 & $-$3.48 & & 10.83 & 6.26 & 0.58 & $-$3.81 & & 17.35 & 5.90 & 0.45 & $-$2.69 \\
ConvLSTM~\cite{shi2015convolutional}  & 11.94 & 4.96 & 0.68 & $-$1.14 & &  9.18 & 4.96 & 0.72 & $-$1.04 & & 16.80 & 4.95 & 0.58 & $-$1.39 \\
SimVP~\cite{gao2022simvp}             & 11.91 & 4.97 & 0.70 & $-$1.14 & &  9.14 & 4.97 & 0.73 & $-$1.05 & & 16.79 & 4.97 & 0.60 & $-$1.37 \\
Earthformer~\cite{gao2022earthformer} & 11.91 & 5.02 & 0.69 & $-$1.18 & &  9.16 & 5.03 & 0.72 & $-$1.12 & & 16.75 & 5.00 & 0.59 & $-$1.32 \\
ClimaX~\cite{nguyen2023climax}        & 11.97 & 5.02 & 0.67 & $-$1.13 & &  9.22 & 5.01 & 0.71 & $-$1.00 & & 16.79 & 5.04 & 0.57 & $-$1.44 \\
TopoFlow~\cite{topoflow}              & 12.01 & 4.96 & 0.71 & $-$1.49 & &  9.21 & 4.95 & 0.74 & $-$1.38 & & 16.91 & 4.98 & 0.62 & $-$1.77 \\
\midrule
\rowcolor{oursrow}
\textbf{CRAN-PM (ours)} & \textbf{11.75} & \textbf{4.84} & \textbf{0.78} & $\mathbf{-}$\textbf{0.86} & & \textbf{8.94} & \textbf{4.86} & \textbf{0.80} & $\mathbf{-}$\textbf{0.75} & & \textbf{16.67} & \textbf{4.78} & \textbf{0.75} & $\mathbf{-}$\textbf{1.13} \\
\bottomrule
\end{tabular}}
\end{table}

CRAN-PM achieves
the best metrics across all station categories. The improvement is most pronounced for
complex terrain ($\sigma_z \geq 50$\,m): bias reduced from $-$1.77 (TopoFlow) to $-$1.13,
a 36\% reduction, demonstrating the effectiveness of elevation-aware attention.

\begin{figure}[!ht]
  \centering
  \includegraphics[width=\textwidth]{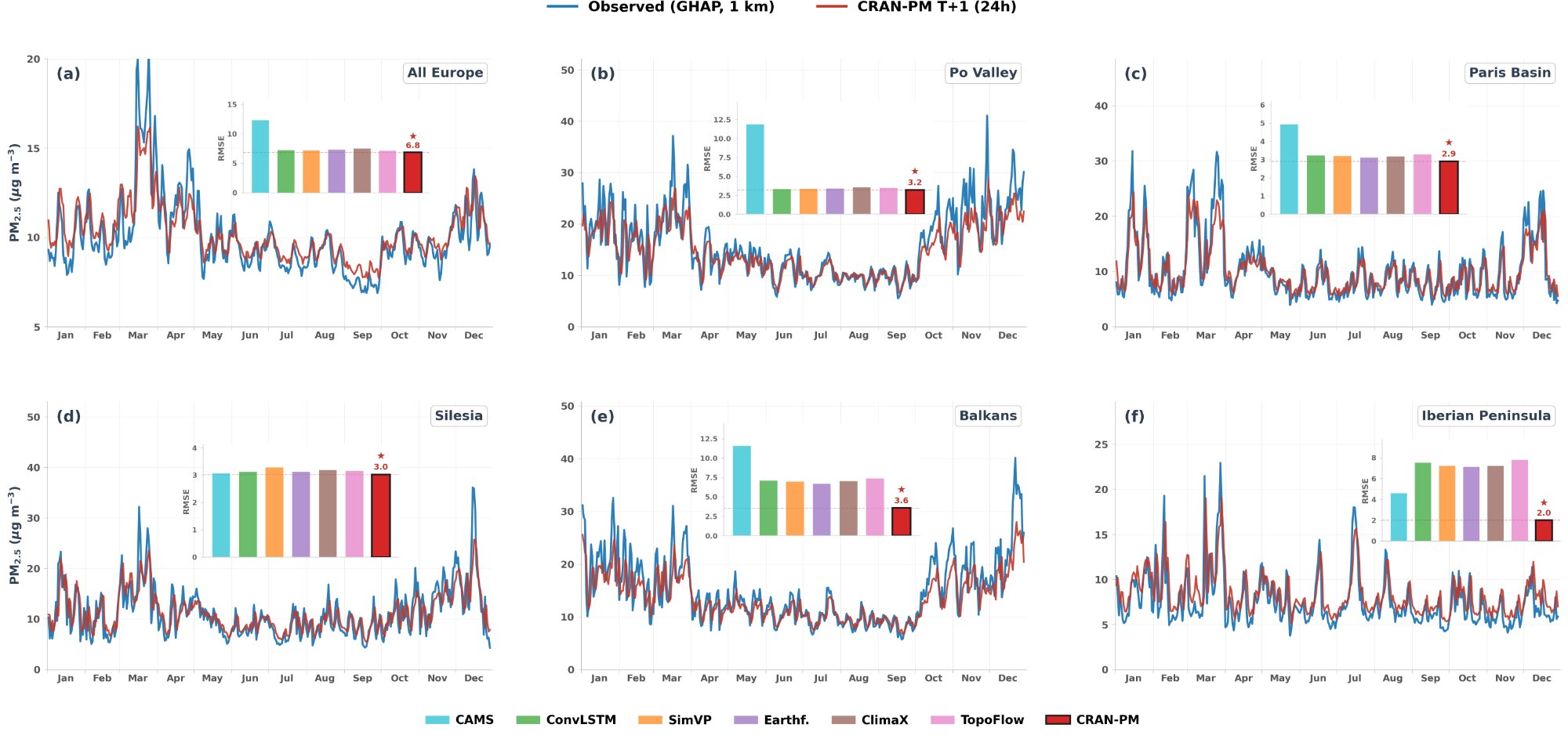}
  \caption{\textbf{Temporal PM$_{2.5}$ evolution (2022, T+1).}
  Blue: GHAP; red: CRAN-PM. Six regions.
  CRAN-PM achieves lowest RMSE in all regions.}
  \label{fig:temporal}
\end{figure}

Figure~\ref{fig:temporal} reveals that CRAN-PM closely tracks the observed
PM$_{2.5}$ seasonal cycle across all six European regions throughout the 2022 test
year. Winter peaks driven by residential heating and temperature inversions are
accurately reproduced, as are summer minima associated with increased boundary
layer height and enhanced precipitation scavenging. The inset RMSE bar plots
confirm that CRAN-PM achieves the lowest error in every region, with the largest
absolute improvements in high-pollution areas such as the Po Valley and Silesia,
where complex orographic confinement amplifies PM$_{2.5}$ accumulation events.

CRAN-PM accurately captures episodic events, including spring biomass
burning plumes in the Balkans and persistent winter smog events in central Europe.
The wind-guided patch reordering and elevation-aware attention are particularly
beneficial for these events: by aligning token sequences with the local advection
field, the model anticipates the spatial direction of plume transport rather than
relying solely on persistence. Degradation from T+1 to T+3
(Table~\ref{tab:main}) indicates that the learned atmospheric dynamics remain
physically consistent over the 72-hour forecast horizon, without spurious error
accumulation across lead times.

Regional variability in RMSE reflects underlying differences in emission density and
meteorological complexity. The Iberian Peninsula records the lowest CRAN-PM RMSE
among all six regions ($2.0\,\mu$g\,m$^{-3}$, Fig.~\ref{fig:temporal}f), consistent
with its predominantly marine boundary layer, low industrial emission density, and
frequent Atlantic ventilation that suppresses PM$_{2.5}$ accumulation. Paris Basin
follows closely at $2.9\,\mu$g\,m$^{-3}$ (panel c). The Balkans present the highest
regional error at $3.6\,\mu$g\,m$^{-3}$ (panel e), driven by episodic wildfire
plumes from outside the training region and steep orographic gradients between the
Dinaric Alps and the Adriatic coast. Po Valley ($3.2\,\mu$g\,m$^{-3}$) and Silesia
($3.0\,\mu$g\,m$^{-3}$) also show elevated errors attributable to intense industrial
sources and thermally stable basin geometries that concentrate PM$_{2.5}$ at
sub-kilometre scales not fully resolved by ERA5 boundary layer parameterisations.
These fine-scale emission hotspots represent the primary source of irreducible
uncertainty in the current architecture.

\begin{figure}[!ht]
  \centering
  \includegraphics[width=\textwidth]{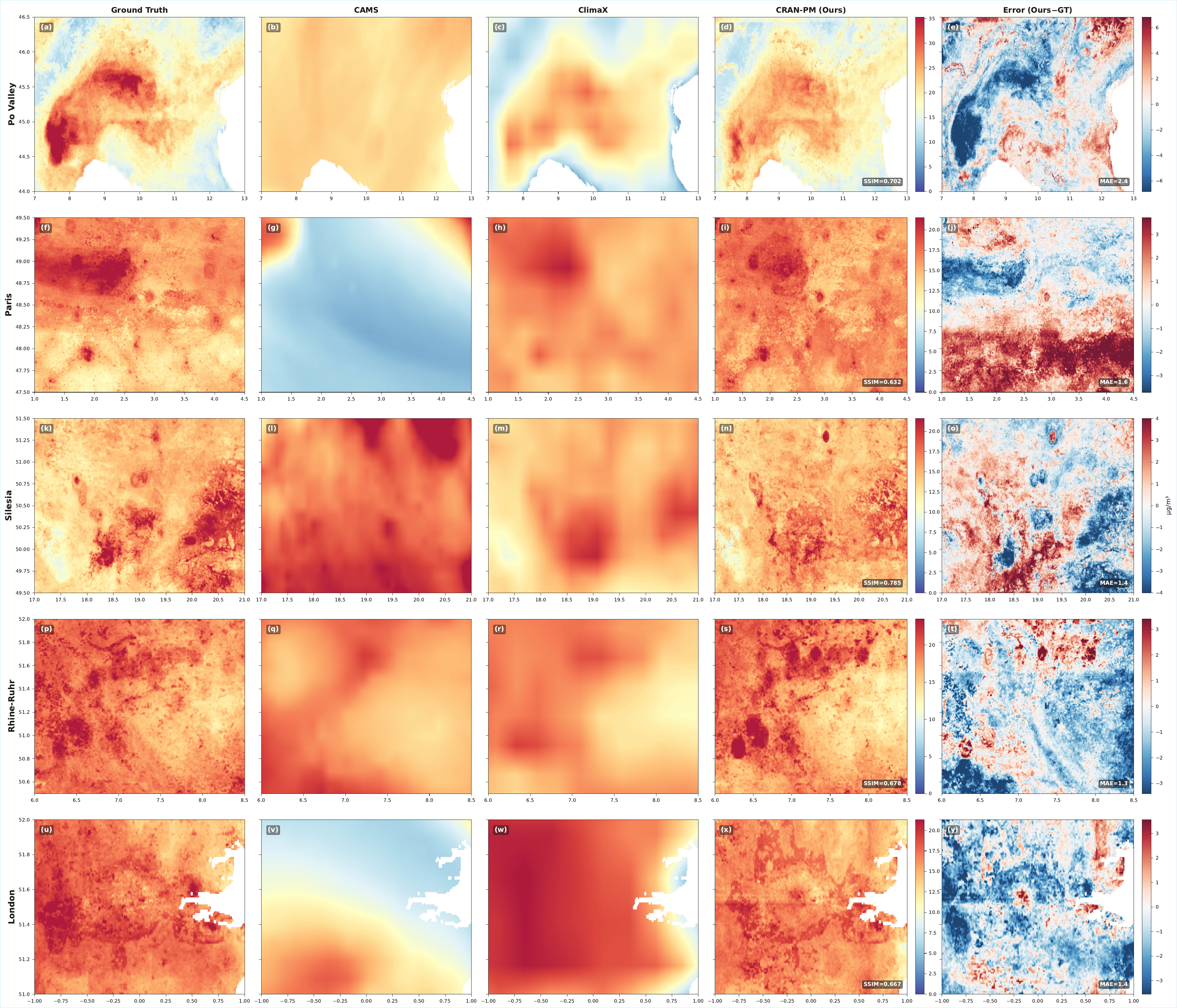}
  \caption{\textbf{Regional PM$_{2.5}$ comparison at 1\,km (T+1).}
  Rows: Po Valley, Paris, Silesia, Rhine-Ruhr, London.
  (a)~GT, (b)~CAMS, (c)~ClimaX, (d)~CRAN-PM, (e)~error (SSIM\,$\geq$\,0.63).}
  \label{fig:spatial}
\end{figure}

\FloatBarrier
\subsection{Ablation Study}
\label{sec:ablation}

\begin{table}[!ht]
\centering
\setlength{\tabcolsep}{4pt}
\caption{\textbf{Ablation study.}
(a)~Cross-attention fusion strategy. (b)~Progressive improvements. (c)~Removal study.
Val RMSE (\textmu g/m$^3$) at T+1, 2022 European test set.}
\label{tab:ablation}
\resizebox{\textwidth}{!}{%
\begin{tabular}{@{} c l cc cccccc ccc r r @{}}
\toprule
& & & &
\rotatebox{70}{Coarse Branch} &
\rotatebox{70}{Cross-Scale Attn} &
\rotatebox{70}{Elev.\ Bias} &
\rotatebox{70}{Wind Scan/Bias} &
\rotatebox{70}{PixelShuffle Dec.} &
\rotatebox{70}{Delta Prediction} &
\rotatebox{70}{FFL\,($\lambda\!=\!0.1$)} &
\rotatebox{70}{Station Loss\,($\lambda\!=\!0.1$)} &
\rotatebox{70}{Day$-$1 Context} &
Val RMSE$\downarrow$ & $\Delta$ \\
\midrule
\multicolumn{15}{@{}l}{\textit{(a) Cross-attention fusion design}} \\
\midrule
& & Q & K/V & \multicolumn{9}{c}{\textit{all other components fixed}} & & \\
\cmidrule(lr){3-4}
(i)    & No cross-attention              & --     & --      & \multicolumn{9}{c}{} & 5.102 & \textcolor{loss}{$+$0.150} \\
(ii)   & Feature addition                & \multicolumn{2}{c}{Sum}    & \multicolumn{9}{c}{} & 5.021 & \textcolor{loss}{$+$0.069} \\
(iii)  & FiLM conditioning               & \multicolumn{2}{c}{Modul.} & \multicolumn{9}{c}{} & 5.036 & \textcolor{loss}{$+$0.084} \\
(iv)   & Concat + Self-Attn              & \multicolumn{2}{c}{Joint}  & \multicolumn{9}{c}{} & 5.040 & \textcolor{loss}{$+$0.088} \\
(v)    & Coarse queries Fine             & Coarse & Fine    & \multicolumn{9}{c}{} & 5.034 & \textcolor{loss}{$+$0.082} \\
(vi)   & Bidirectional                   & Both   & Both    & \multicolumn{9}{c}{} & 5.015 & \textcolor{loss}{$+$0.063} \\
\textbf{(vii)} & \textbf{Cross-resolution attention (ours)} & \textbf{Fine} & \textbf{Coarse} & \multicolumn{9}{c}{} & \textbf{4.952} & \textbf{ref.} \\
\midrule
\multicolumn{15}{@{}l}{\textit{(b) Progressive improvements}} \\
\midrule
(A) & Baseline (MSE only)        & Fine & Coarse &\cmark&\cmark&\cmark&\cmark&\cmark&\cmark&\xmark&\xmark&\xmark & 5.24 & \textcolor{gray}{ref.} \\
(B) & + Eff.\ batch size         & Fine & Coarse &\cmark&\cmark&\cmark&\cmark&\cmark&\cmark&\xmark&\xmark&\xmark & 5.10 & \textcolor{gain}{$-$0.14} \\
(C) & + Focal Frequency Loss     & Fine & Coarse &\cmark&\cmark&\cmark&\cmark&\cmark&\cmark&\cmark&\xmark&\xmark & 5.04 & \textcolor{gain}{$-$0.06} \\
(D) & + Station loss \& day$-$1  & Fine & Coarse &\cmark&\cmark&\cmark&\cmark&\cmark&\cmark&\cmark&\cmark&\cmark & 4.95 & \textcolor{gain}{$-$0.09} \\
\midrule
\multicolumn{15}{@{}l}{\textit{(c) Removal study (from full CRAN-PM)}} \\
\midrule
 & \textbf{CRAN-PM (full)} &\textbf{Fine}&\textbf{Coarse}&\cmark&\cmark&\cmark&\cmark&\cmark&\cmark&\cmark&\cmark&\cmark & \textbf{4.95} & \textbf{ref.} \\
\midrule
(R1) & $-$ Cross-scale attention    & --   & --     &\cmark&\xmark&\cmark&\cmark&\cmark&\cmark&\cmark&\cmark&\cmark & 5.38 & \textcolor{loss}{$+$0.43} \\
(R2) & $-$ Elevation bias           & Fine & Coarse &\cmark&\cmark&\xmark&\cmark&\cmark&\cmark&\cmark&\cmark&\cmark & 5.12 & \textcolor{loss}{$+$0.17} \\
(R3) & $-$ Wind scan \& bias        & Fine & Coarse &\cmark&\cmark&\cmark&\xmark&\cmark&\cmark&\cmark&\cmark&\cmark & 5.09 & \textcolor{loss}{$+$0.14} \\
(R4) & $-$ PixelShuffle (ConvT)     & Fine & Coarse &\cmark&\cmark&\cmark&\cmark&\xmark&\cmark&\cmark&\cmark&\cmark & 5.08 & \textcolor{loss}{$+$0.13} \\
(R5) & $-$ Delta prediction         & Fine & Coarse &\cmark&\cmark&\cmark&\cmark&\cmark&\xmark&\cmark&\cmark&\cmark & 5.18 & \textcolor{loss}{$+$0.23} \\
(R6) & $-$ FFL                      & Fine & Coarse &\cmark&\cmark&\cmark&\cmark&\cmark&\cmark&\xmark&\cmark&\cmark & 5.07 & \textcolor{loss}{$+$0.12} \\
(R7) & $-$ Station loss             & Fine & Coarse &\cmark&\cmark&\cmark&\cmark&\cmark&\cmark&\cmark&\xmark&\cmark & 5.02 & \textcolor{loss}{$+$0.07} \\
(R8) & $-$ Day$-$1 context          & Fine & Coarse &\cmark&\cmark&\cmark&\cmark&\cmark&\cmark&\cmark&\cmark&\xmark & 5.04 & \textcolor{loss}{$+$0.09} \\
\bottomrule
\end{tabular}}
\end{table}

\paragraph{(a) Cross-attention fusion.}
Cross-resolution attention (row~vii, 4.952) outperforms all alternatives. Removing
cross-attention (row~i, $+$0.150) is the largest single degradation. The reverse
direction (row~v, $+$0.082) validates the physical asymmetry: local PM$_{2.5}$
responds to large-scale meteorology, not vice versa.

\paragraph{(b) Progressive improvements.}
Larger effective batch ($-$0.14) stabilises training across heterogeneous tiles.
Focal frequency loss ($-$0.06) sharpens gradients at urban-rural boundaries. Station
loss and day$-$1 context ($-$0.09) correct satellite retrieval biases and provide
atmospheric tendency.

\paragraph{(c) Removal study.}
Cross-scale attention ($+$0.43) and delta prediction ($+$0.23) are the two most
critical components. Physics-guided biases jointly contribute $+$0.31 (R2+R3).

Table~\ref{tab:ablation} shows a clear ranking of components: the
cross-resolution attention bridge provides the single largest performance gain,
confirming that fusing coarse meteorological context with fine-scale observations
is the core architectural innovation. The elevation bias (R2, $+$0.17) and wind
scan (R3, $+$0.14) contribute comparably, reflecting the physical importance of
orographic confinement and advection for PM$_{2.5}$ accumulation over complex
terrain. Delta prediction (R5, $+$0.23) is the second most critical component,
as initialising the network output to today's PM$_{2.5}$ field
reduces the learning burden: the model only needs to predict the daily change
rather than the absolute concentration, which eases optimisation
for a residual mapping near zero over clean-air days.

The progressive improvement experiment (Table~\ref{tab:ablation}b) demonstrates
that each training innovation is independently beneficial and their combination
is approximately additive. The focal frequency loss targets a well-known weakness
of $\ell_2$-based objectives: the tendency to over-smooth sharp spatial
gradients at emission source boundaries. Station loss ($\lambda_\text{station} =
0.1$) anchors predictions to ground-measured concentrations,
reducing the systematic negative bias inherited from GHAP satellite
retrievals in complex terrain. The day$-$1 context input adds an
atmospheric tendency that helps the model capture short-term trends
without explicit recurrence.

\begin{figure}[!ht]
  \centering
  \includegraphics[width=\textwidth]{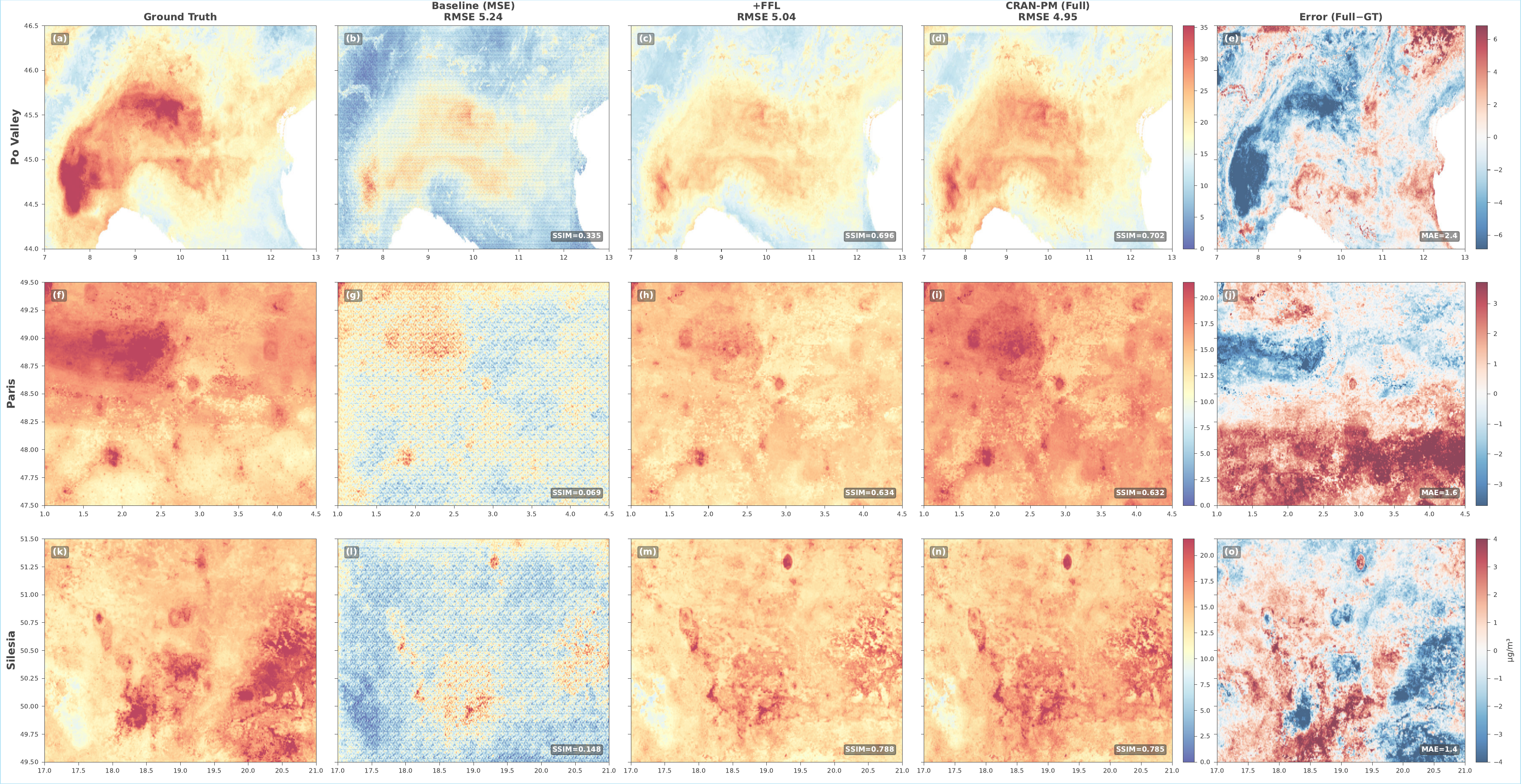}
  \caption{\textbf{Ablation: progressive improvements at 1\,km (T+1).}
  (a)~GT, (b)~baseline (SSIM\,$\leq$\,0.34), (c)~+FFL,
  (d)~CRAN-PM full (RMSE\,=\,4.95), (e)~error.}
  \label{fig:ablation_spatial}
\end{figure}

\FloatBarrier
\noindent\textbf{Limitations and Discussion}While CRAN-PM achieves strong results, several directions remain open. The model relies on GHAP satellite-derived PM$_{2.5}$ as supervision, which carries retrieval uncertainties under clouds and complex terrain. The current temporal resolution is daily, whereas operational air quality management often benefits from hourly forecasts. The prediction horizon is limited to 1$\sim$3 days; extending it would require modelling non-stationary emissions and atmospheric chemistry. Finally, the fixed $512\times512$ tiling strategy could be improved with adaptive spatial partitioning.


\section{Conclusion}
\label{sec:conclusion}

We introduced \textbf{CRAN-PM}, a dual-branch Vision Transformer for ultra-high-resolution PM$_{2.5}$ prediction. The key contribution is cross-resolution attention, linking global coarse-resolution meteorology with local high-resolution PM$_{2.5}$ observation. The global branch captures large-scale context, while local branches model fine-scale details, and the cross-attention enables efficient information exchange across scales. This design preserves long-range dependencies while reducing peak memory from hundreds of gigabytes to under 2\,GB per tile, with computation scaling linearly with domain size. Physics-guided attention biases, elevation-aware self-attention and wind-guided cross-attention, inject domain knowledge as soft priors for physical regularization, allowing the network to produce physically consistent prediction. On daily PM$_{2.5}$ forecasting across Europe (2022), CRAN-PM achieves RMSE of 6.85\,\textmu g/m$^3$ at 1\,km (T+1), improving 4.7$\sim$10.7\% over the strongest single-scale baseline and reducing bias at complex-terrain stations by 36\%. More generally, cross-resolution attention offers a scalable mechanism for Vision Transformers in ultra-high-resolution tasks requiring global context, with potential applications in urban-scale environmental modeling and analysis.

\printbibliography[heading=bibintoc,title=References]

@inproceedings{kheder2025aqnet,
  title     = {Deep Spatio-Temporal Neural Network for Air Quality Reanalysis},
  author    = {Kheder, Ammar and Foreback, Benjamin and Wang, Lili and Liu, Zhi-Song and Boy, Michael},
  booktitle = {Image Analysis: 23rd Scandinavian Conference, SCIA 2025},
  series    = {Lecture Notes in Computer Science},
  volume    = {15725},
  publisher = {Springer},
  year      = {2025}
}

@inproceedings{vptr,title={VPTR: Efficient Transformers for Video Prediction},author={Ye, Xi and Bilodeau, Guillaume-Alexandre},booktitle={ICPR},pages={3492--3499},year={2022},organization={IEEE}}

@INPROCEEDINGS{vivit,author={Arnab, Anurag and Dehghani, Mostafa and Heigold, Georg and Sun, Chen and Lu\v{c}i\'{c}, Mario and Schmid, Cordelia},booktitle={ICCV},title={ViViT: A Video Vision Transformer},year={2021},pages={6816-6826}}

@article{spinode,title={SPIN-ODE},author={Wenqing Peng and Zhi-Song Liu and Michael Boy},year={2025},journal={ECAI},pages={2033--2040}}

@article{nne,title={Neural network emulator for atmospheric chemical ODE},journal={Neural Networks},volume={184},pages={107106},year={2025},author={Zhi-Song Liu and Petri Clusius and Michael Boy}}

@article{topoflow,title={TopoFlow: Physics-guided Neural Networks for High-Resolution Air Quality Prediction},author={Kheder, Ammar and Toropainen, Helmi and Peng, Wenqing and Ant{\~a}o, Samuel and Chen, Jia and Liu, Zhi-Song and Boy, Michael},journal={arXiv preprint arXiv:2602.16821},year={2026}}

@article{bi2023pangu,title={Accurate medium-range global weather forecasting with {3D} neural networks},author={Bi, Kaifeng and Xie, Lingxi and Zhang, Hengheng and Chen, Xin and Gu, Xiaotao and Tian, Qi},journal={Nature},volume={619},pages={533--538},year={2023},publisher={Nature Publishing Group}}

@article{lam2023graphcast,title={Learning skillful medium-range global weather forecasting},author={Lam, Remi and Sanchez-Gonzalez, Alvaro and Willson, Matthew and others},journal={Science},volume={382},number={6677},pages={1416--1421},year={2023}}

@article{bodnar2024aurora,title={Aurora: A Foundation Model of the Atmosphere},author={Bodnar, Cristian and Bruinsma, Wessel P and others},journal={Nature},year={2024}}

@inproceedings{nguyen2023climax,title={{ClimaX}: A Foundation Model for Weather and Climate},author={Nguyen, Tung and Brandstetter, Johannes and Kapoor, Ashish and Gupta, Jayesh K and Grover, Aditya},booktitle={ICML},year={2023}}

@article{pathak2022fourcastnet,title={{FourCastNet}},author={Pathak, Jaideep and others},journal={arXiv preprint arXiv:2202.11214},year={2022}}

@article{wei2023ghap,title={Reconstructing 1-km-resolution high-quality {PM}$_{2.5}$ data records},author={Wei, Jing and Li, Zhanqing and Lyapustin, Alexei and others},journal={Remote Sensing of Environment},volume={252},pages={112136},year={2021}}

@article{cams,title={The {CAMS} reanalysis of atmospheric composition},author={Inness, Antje and Ades, Melanie and others},journal={Atmospheric Chemistry and Physics},volume={19},number={6},pages={3515--3556},year={2019}}

@article{chen2019xgboost,title={Extreme Gradient Boosting model to estimate {PM}$_{2.5}$},author={Chen, Gang and others},journal={Atmospheric Environment},volume={202},pages={180--189},year={2019}}

@article{wang2020pm25gnn,title={{PM}$_{2.5}$-{GNN}},author={Wang, Shuo and others},journal={arXiv preprint arXiv:2002.12898},year={2020}}

@inproceedings{shi2015convolutional,title={Convolutional {LSTM} Network},author={Shi, Xingjian and Chen, Zhourong and Wang, Hao and others},booktitle={NeurIPS},year={2015}}

@inproceedings{gao2022simvp,title={{SimVP}: Simpler yet Better Video Prediction},author={Gao, Zhangyang and Tan, Cheng and Wu, Lirong and Li, Stan Z},booktitle={CVPR},year={2022}}

@inproceedings{gao2022earthformer,title={Earthformer},author={Gao, Zhihan and Shi, Xingjian and Wang, Hao and others},booktitle={NeurIPS},year={2022}}

@article{raissi2019physics,title={Physics-informed neural networks},author={Raissi, Maziar and Perdikaris, Paris and Karniadakis, George E},journal={Journal of Computational Physics},volume={378},pages={686--707},year={2019}}

@inproceedings{guen2020phydnet,title={Disentangling Physical Dynamics from Unknown Factors},author={Le Guen, Vincent and Thome, Nicolas},booktitle={CVPR},pages={11474--11484},year={2020}}

@article{zhang2023nowcastnet,title={Skilful nowcasting of extreme precipitation with {NowcastNet}},author={Zhang, Yuchen and Long, Mingsheng and others},journal={Nature},volume={619},pages={526--532},year={2023}}

@inproceedings{shi2016pixelshuffle,title={Real-Time Single Image and Video Super-Resolution Using an Efficient Sub-Pixel Convolutional Neural Network},author={Shi, Wenzhe and Caballero, Jose and others},booktitle={CVPR},pages={1874--1883},year={2016}}

@inproceedings{liu2021swin,title={Swin {Transformer}: Hierarchical Vision Transformer Using Shifted Windows},author={Liu, Ze and Lin, Yutong and Cao, Yue and others},booktitle={ICCV},pages={10012--10022},year={2021}}

@inproceedings{jiang2021focal,title={Focal Frequency Loss for Image Reconstruction and Synthesis},author={Jiang, Liming and Dai, Bo and Wu, Wayne and Loy, Chen Change},booktitle={ICCV},pages={13919--13929},year={2021}}

@book{whiteman2000mountain,title={Mountain Meteorology: Fundamentals and Applications},author={Whiteman, C David},year={2000},publisher={Oxford University Press}}

@inproceedings{dosovitskiy2021image,title={An Image is Worth 16x16 Words: Transformers for Image Recognition at Scale},author={Dosovitskiy, Alexey and Beyer, Lucas and Kolesnikov, Alexander and others},booktitle={ICLR},year={2021}}

@inproceedings{liu2022video,title={Video Swin Transformer},author={Liu, Ze and Ning, Jia and Cao, Yue and others},booktitle={CVPR},year={2022}}

@inproceedings{bertasius2021spacetime,title={Is Space-Time Attention All You Need for Video Understanding?},author={Bertasius, Gedas and Wang, Heng and Torresani, Lorenzo},booktitle={ICML},year={2021}}

@inproceedings{chen2022hipt,title={Scaling Vision Transformers to Gigapixel Images via Hierarchical Self-Supervised Learning},author={Chen, Richard J. and others},booktitle={CVPR},year={2022}}

@inproceedings{wang2021pvt,title={Pyramid Vision Transformer},author={Wang, Wenhai and others},booktitle={ICCV},year={2021}}

@inproceedings{chu2021twins,title={Twins: Revisiting the Design of Spatial Attention in Vision Transformers},author={Chu, Xiangxiang and others},booktitle={NeurIPS},year={2021}}

@inproceedings{chen2021crossvit,title={CrossViT: Cross-Attention Multi-Scale Vision Transformer for Image Classification},author={Chen, Chun-Fu and Fan, Quanfu and Panda, Rameswar},booktitle={ICCV},year={2021}}

@inproceedings{shao2021transmil,title={TransMIL},author={Shao, Zhuchen and others},booktitle={NeurIPS},year={2021}}

@article{chen2023fuxi,title={FuXi},author={Chen, Lei and others},journal={npj Climate and Atmospheric Science},year={2023}}

@article{price2024gencast,title={GenCast},author={Price, Ilan and others},journal={Nature},year={2024}}

@article{li2021fno,title={Fourier Neural Operator},author={Li, Zongyi and others},journal={ICLR},year={2021}}

@article{li2024pino,title={Physics-Informed Neural Operator},author={Li, Zongyi and others},journal={JMLR},year={2024}}

\clearpage
\newpage
\beginsupplement

\begin{center}
  {\Large\textbf{Supplementary Material}}\\[6pt]
  {\normalsize Cross-Resolution Attention Network for High-Resolution PM$_{2.5}$ Prediction}
\end{center}
\vspace{4pt}

\noindent This supplementary material provides: (1) a scalability analysis motivating the
dual-branch design, (2) zero-shot geographic transferability results on two
out-of-distribution regions, and (3) complete architecture specifications,
hyperparameters, test set stratification, input channels, and the forward pass algorithm.

\section{Architecture Overview and Scalability Motivation}
\label{sec:s_overview}

\begin{figure}[!ht]
  \centering
  \includegraphics[width=\textwidth]{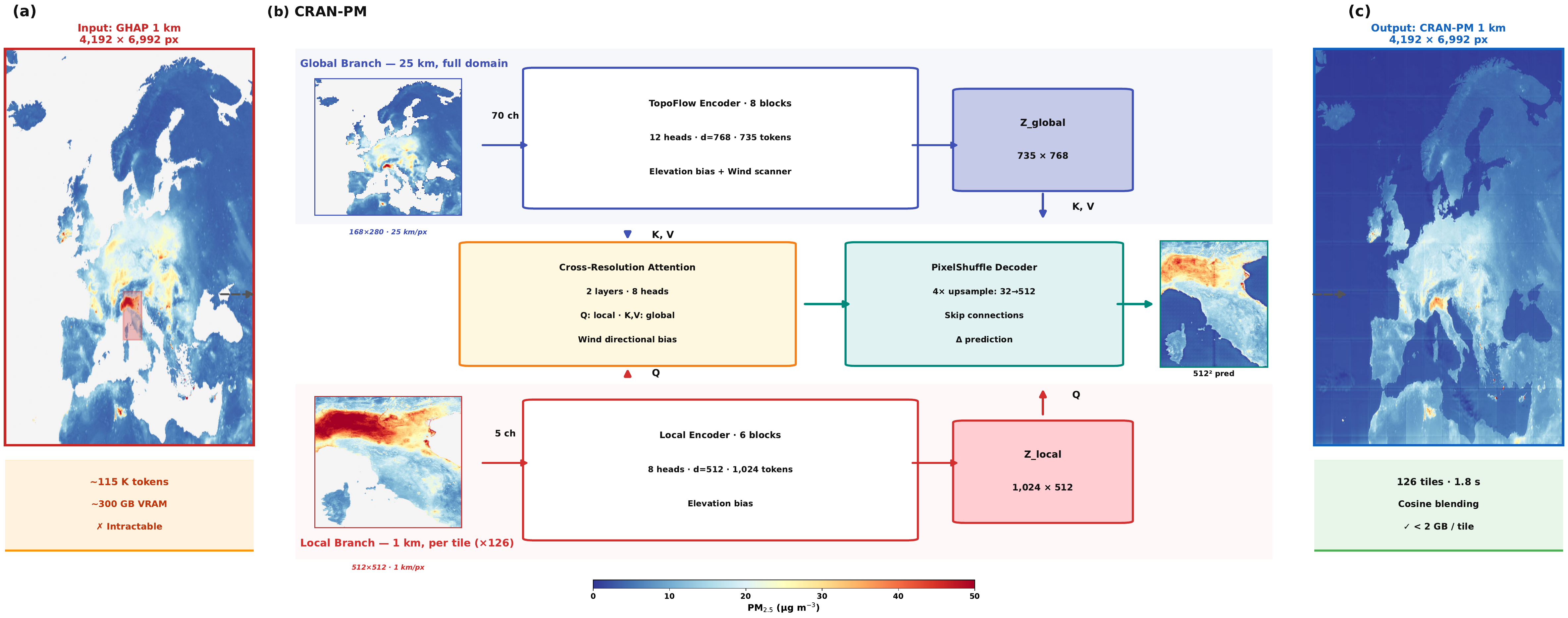}
  \caption{\textbf{CRAN-PM architecture overview.}
  \textit{Left}: naive 1\,km processing is intractable ($\sim$115\,K tokens,
  $\sim$300\,GB VRAM).
  \textit{Centre}: dual-branch design with Global Branch (735 tokens, 25\,km) and Local
  Branch (1{,}024 tokens/tile, 1\,km), fused by Cross-Resolution Attention with wind bias.
  \textit{Right}: full-domain output from 126 overlapping tiles with cosine blending
  ($<$2\,GB/tile, $\sim$1.8\,s).}
  \label{fig:arch_overview}
\end{figure}

At 1\,km resolution, the full European domain yields \textbf{115\,K tokens} requiring
$\sim$\textbf{300\,GB VRAM}, intractable due to $\mathcal{O}(N^2)$ attention.
CRAN-PM solves this via scale decoupling (Fig.~\ref{fig:arch_overview}): a
\textbf{Global Branch} encodes ERA5/CAMS (70\,ch, 25\,km) into $\mathbf{Z}_\text{global}$
once per day, while a \textbf{Local Branch} processes each $512\!\times\!512$ GHAP tile
independently, fused by \textbf{Cross-Resolution Attention} ($Q$: local, $KV$: global,
wind bias $\mathbf{B}^\text{wind}$), yielding $<$\textbf{2\,GB/tile} at
$\sim$\textbf{1.8\,s} per inference.

The tile-based inference strategy mitigates potential boundary artefacts by
overlapping adjacent tiles by 64\,pixels and applying a cosine blending weight that
smoothly down-weights predictions near tile edges, yielding a seamless full-domain
output. The shared Global Branch with cached daily representations ($735 \times
735$-token attention computed once per day and reused across all 126 local tiles)
reduces total inference time by approximately 60\% compared to a non-cached
architecture, enabling full-domain PM$_{2.5}$ forecasts in under 4\,minutes on a
single AMD MI250X GPU.

Memory scaling is a key advantage of the tile-based approach. At inference,
peak GPU memory per tile is under 2\,GB, enabling deployment on any GPU with
at least 4\,GB HBM. At training time, a per-GPU batch of 2 tiles at bfloat16
precision fits within the 64\,GB HBM of each AMD MI250X GCD. The daily caching
of the Global Branch further reduces training memory: only the Local Branch and
Cross-Resolution Attention are backpropagated per tile, while the Global Branch
gradient is accumulated once per day over all 126 tiles sharing the same
meteorological forcing.

\section{Out-of-Distribution Generalisation}
\label{sec:s_ood}

To assess geographic transferability, we apply the Europe-trained checkpoint zero-shot
to two out-of-distribution regions using the same ERA5/CAMS/GHAP pipeline, with no
domain adaptation or fine-tuning.

\textbf{North America} shares similar pollution regimes with Europe: wildfire-driven
PM$_{2.5}$, comparable concentration ranges (0--20\,\textmu g/m$^3$), and analogous
orographic patterns (Rocky Mountains vs.\ European Alps), placing test samples largely
within the training distribution. \textbf{India}, by contrast, presents a fundamentally
different regime: the Indo-Gangetic Plain (IGP) experiences extreme anthropogenic
pollution (up to 80\,\textmu g/m$^3$, i.e.\ 3--5$\times$ the European training range),
with emission sources (brick kilns, crop burning, dense vehicle traffic) that have no
European equivalent.

\begin{figure}[!ht]
  \centering
  \includegraphics[width=\textwidth]{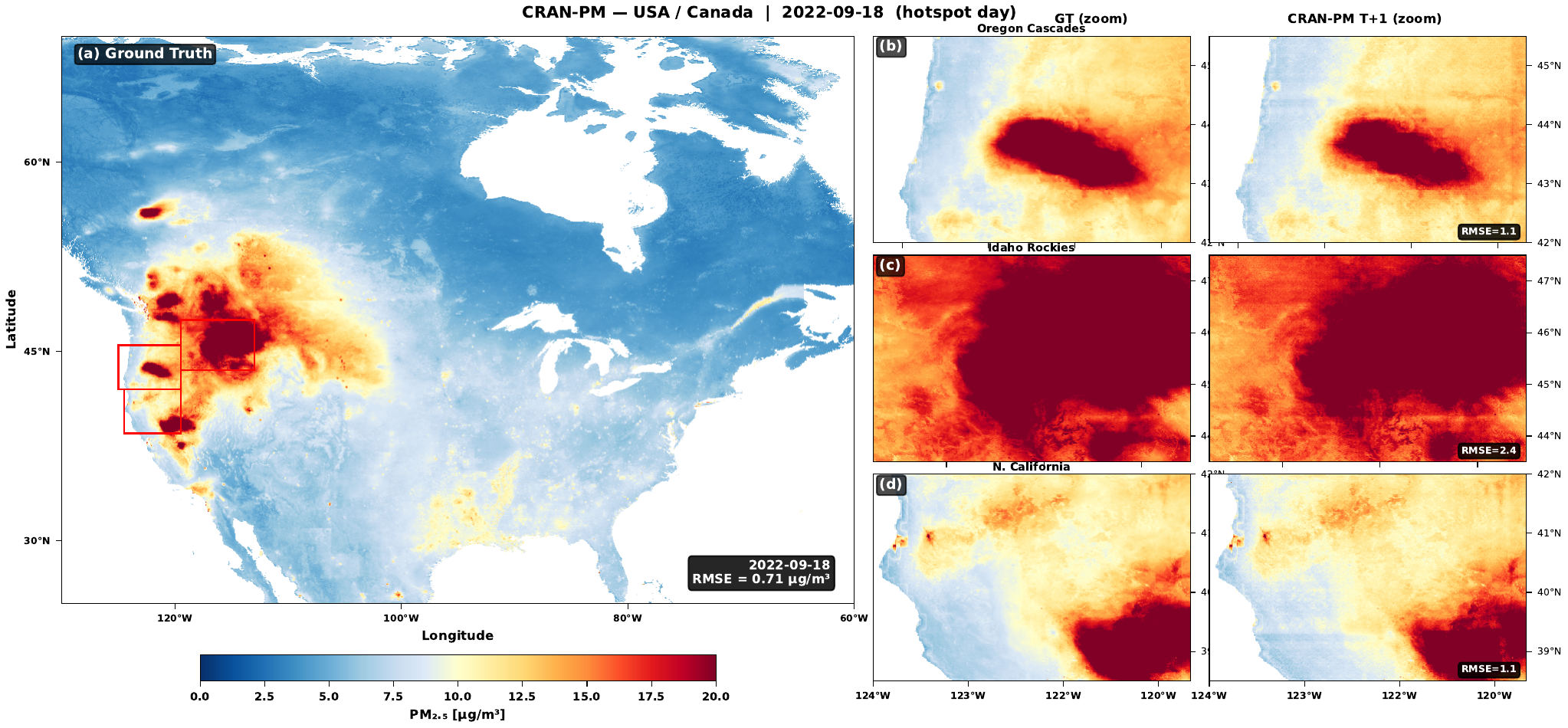}
  \caption{\textbf{Zero-shot transfer to USA/Canada} (2022-09-18, wildfire episode).
    RMSE\,=\,0.71\,\textmu g/m$^3$ overall; Oregon Cascades\,=\,1.1,
    Idaho Rockies\,=\,2.4, N.\,California\,=\,1.1.}
  \label{fig:ood_usa}
\end{figure}

\begin{figure}[!ht]
  \centering
  \includegraphics[width=\textwidth]{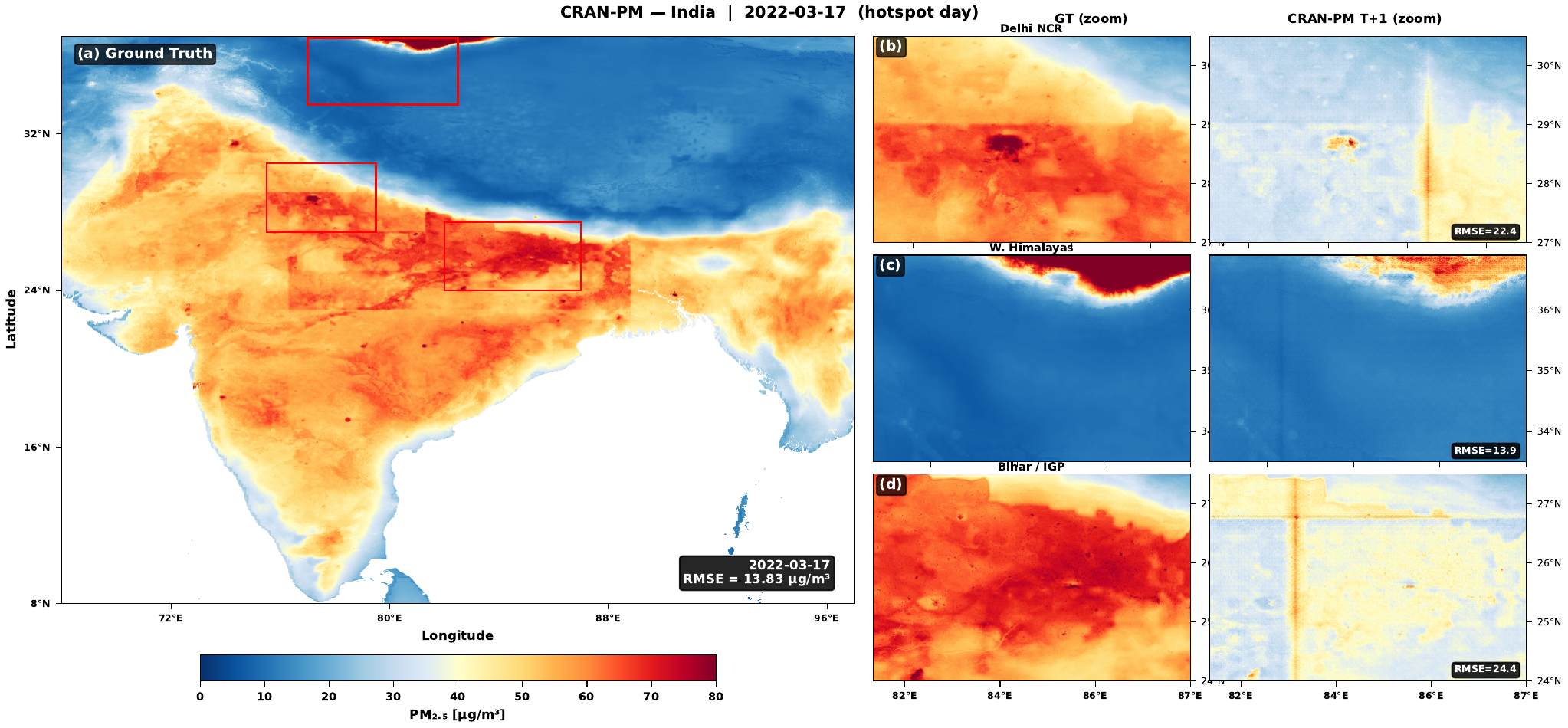}
  \caption{\textbf{Zero-shot transfer to India} (2022-03-17, IGP hotspot day).
    RMSE\,=\,13.83\,\textmu g/m$^3$ overall; Delhi NCR\,=\,22.4, Bihar/IGP\,=\,24.4.
    IGP concentrations (up to 80\,\textmu g/m$^3$) are 3--5$\times$ the European
    training range.}
  \label{fig:ood_india}
\end{figure}

\begin{figure}[!ht]
  \centering
  \begin{minipage}[t]{0.48\textwidth}
    \centering
    \includegraphics[width=\textwidth]{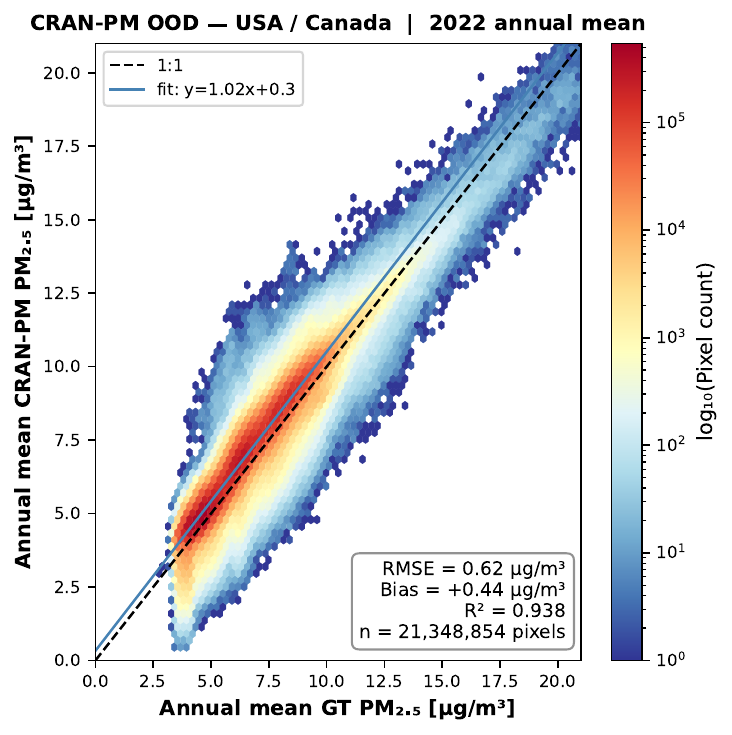}
  \end{minipage}
  \hfill
  \begin{minipage}[t]{0.48\textwidth}
    \centering
    \includegraphics[width=\textwidth]{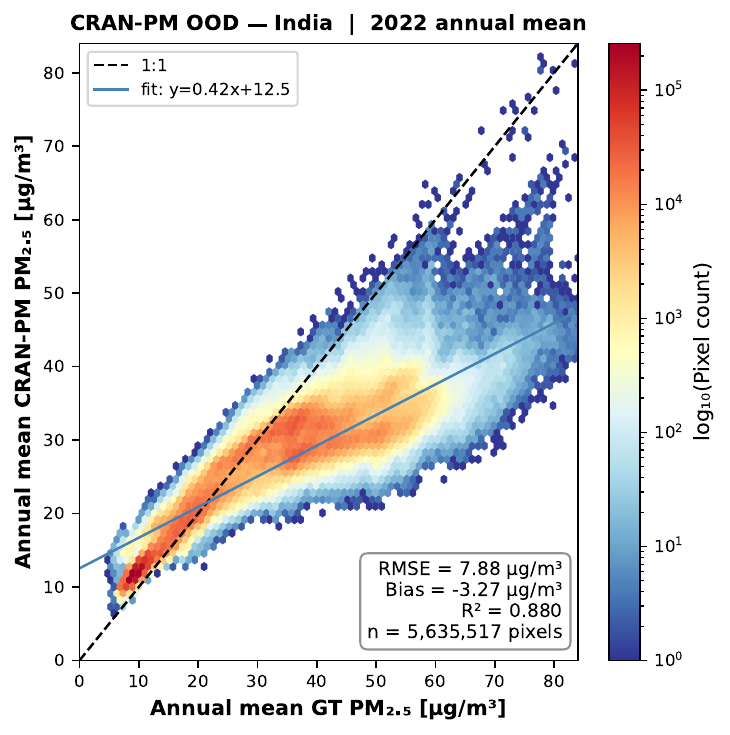}
  \end{minipage}
  \caption{\textbf{Annual mean scatter plots (2022) for both OOD regions.}
  \textit{Left} (USA/Canada): RMSE\,=\,0.62, Bias\,=\,$+$0.44, $R^2$\,=\,0.938
  ($n$\,=\,21.3M px); slope $y\!=\!1.02x+0.3$ confirms near-unity scaling.
  \textit{Right} (India): RMSE\,=\,7.88, Bias\,=\,$-$3.27, $R^2$\,=\,0.880
  ($n$\,=\,5.6M px); systematic underestimation above 40\,\textmu g/m$^3$.}
  \label{fig:ood_scatter}
\end{figure}

For each region we report spatial maps (Fig.~\ref{fig:ood_usa}--\ref{fig:ood_india})
and annual mean scatter plots (Fig.~\ref{fig:ood_scatter}) aggregating 365 daily
predictions over millions of pixels.

The near-unity regression slope on USA/Canada ($y\!=\!1.02x+0.3$, $R^2\!=\!0.938$)
suggests the model has internalised transferable representations of meteorology-driven
PM$_{2.5}$ dynamics rather than memorising European spatial patterns. The India result
confirms that transferability is bounded by the emission regime.

\textbf{Practical implications.}
These zero-shot experiments suggest that CRAN-PM can be deployed with minimal
adaptation in regions sharing European pollution meteorology, such as temperate
North America or western Asia. For regions with structurally distinct emission
regimes, such as the Indo-Gangetic Plain, a lightweight fine-tuning step on a
small set of locally labelled GHAP tiles ($\sim$3--6\,months of data) is expected
to close the performance gap, since the backbone meteorological representations
learned from ERA5 remain broadly transferable across climatic zones. Future work
will explore continual adaptation strategies that allow CRAN-PM to update its
emission-regime priors without retraining the full model.

\section{Architecture and Implementation Details}
\label{sec:s_arch}

CRAN-PM is a 96\,M-parameter dual-branch Vision Transformer. The \textbf{Global Branch}
encodes ERA5/CAMS coarse inputs (70 channels, 25\,km) into a 735-token representation
$\mathbf{Z}^g$ once per forecast day, cached across all 126 tiles. The \textbf{Local
Branch} encodes each $512\!\times\!512$ GHAP tile (5 channels, 1\,km) into 1024 tokens.
A \textbf{Cross-Resolution Attention bridge} fuses the two scales using wind-directional
bias, and a \textbf{PixelShuffle decoder} reconstructs the full-resolution residual
$\Delta$, added back to today's PM$_{2.5}$ tile.

{\small
\renewcommand\arraystretch{1.2}
\setlength{\tabcolsep}{4pt}

\smallskip
\noindent\textbf{Table~S1: Layer-by-layer architecture.}
\smallskip

\captionof{table}{\textbf{CRAN-PM layer-by-layer architecture.}
96\,M parameters; FFN ratio 4.0; drop-path 0.1; input dropout 0.1.}
\label{tab:s_arch}
\resizebox{\textwidth}{!}{%
\begin{tabular}{@{} l l l l l @{}}
\toprule
\textbf{Module} & \textbf{Input} & \textbf{Output} & \textbf{Config} & \textbf{Notes} \\
\midrule
\multicolumn{5}{@{}l}{\textit{Global branch}} \\
Unfold + Linear           & $70\!\times\!168\!\times\!280$ & $735\!\times\!768$ & $k\!=\!s\!=\!8$ & rocBLAS \\
Sinusoidal 2D pos.\ embed & $735\!\times\!768$             & $735\!\times\!768$ & fixed & \\
Elevation-aware Attn      & $735\!\times\!768$             & $735\!\times\!768$ & $h\!=\!12$,\;$E_0\!=\!1000$\,m & 1 block \\
Swin Transformer          & $735\!\times\!768$             & $735\!\times\!768$ & $h\!=\!12$, win $7\!\times\!7$, depth 7 & shifted window \\
\midrule
\multicolumn{5}{@{}l}{\textit{Local branch (per tile)}} \\
Unfold + Linear           & $5\!\times\!512\!\times\!512$  & $1024\!\times\!512$ & $k\!=\!s\!=\!16$ & rocBLAS \\
Sinusoidal 2D pos.\ embed & $1024\!\times\!512$            & $1024\!\times\!512$ & fixed & \\
Lead-time embedding       & $\tau\!\in\!\{1,2,3\}$        & $1024\!\times\!512$ & learnable $\mathbb{R}^{512}$ & broadcast \\
Elevation-aware Attn      & $1024\!\times\!512$            & $1024\!\times\!512$ & $h\!=\!8$,\;$E_0\!=\!500$\,m & 1 block \\
Swin Transformer          & $1024\!\times\!512$            & $1024\!\times\!512$ & $h\!=\!8$, win $8\!\times\!8$, depth 5 & \\
\midrule
\multicolumn{5}{@{}l}{\textit{Cross-resolution bridge}} \\
Linear projection         & $735\!\times\!768$             & $735\!\times\!512$ & no bias & global $\to$ local dim \\
Cross-Attn $\times$2 + FFN & Q:$1024\!\times\!512$, KV:$735\!\times\!512$ & $1024\!\times\!512$ & $h\!=\!8$, head dim 64 & wind bias $\beta$ per head \\
\midrule
\multicolumn{5}{@{}l}{\textit{PixelShuffle decoder}} \\
Reshape                   & $1024\!\times\!512$            & $512\!\times\!32\!\times\!32$ & & \\
Upblock $\times$4 (Conv+PS+Res) &                          & $32\!\times\!512\!\times\!512$ & $r\!=\!2$ each & \\
Conv $1\!\times\!1$, zero-init & $32\!\times\!512$         & $1\!\times\!512$              & & residual $\Delta$ \\
\bottomrule
\end{tabular}}
\medskip

\noindent The Global Branch uses a larger Swin depth (7 blocks) than the Local
Branch (5 blocks) to capture planetary-scale transport patterns. The elevation-aware
attention block is placed before the Swin Transformer in each branch so that
topographically-induced concentration gradients are encoded into the token
representations before long-range spatial mixing. The PixelShuffle decoder
operates at $r\!=\!2$ per upsampling step (4 steps: $32\!\to\!64\!\to\!128\!\to\!256\!\to\!512$
pixels), which empirically outperformed a single-step transpose convolution at
equivalent parameter count, likely because successive upsampling preserves
high-frequency spatial structure more faithfully.

\noindent The test set is stratified to ensure balanced coverage across seasons,
times of day, and days of the month (Table~\ref{tab:s_testset}). Equal sample
counts (1{,}000 per lead time) prevent any single temporal regime from dominating
reported metrics. The balanced seasonal distribution is particularly important
for PM$_{2.5}$ evaluation: winter episodes (higher concentrations, stronger
orographic gradients) and summer conditions (lower background, photochemical
secondary formation) require qualitatively different predictive skills. All 4{,}000
test samples are drawn from 2022, providing a strictly out-of-sample evaluation
with zero temporal leakage from the 2017--2021 training period.

\renewcommand\arraystretch{1.05}
\captionof{table}{\textbf{Test set stratification} (year 2022, 4{,}000 samples).}
\label{tab:s_testset}
\resizebox{\textwidth}{!}{%
\begin{tabular}{c|cccc|cccc|ccc|c}
\toprule
\multirow{2}{*}{\begin{tabular}[c]{@{}c@{}}Temporal\\Horizon\end{tabular}}
  & \multicolumn{4}{c|}{Seasonal Distribution}
  & \multicolumn{4}{c|}{Daily Distribution}
  & \multicolumn{3}{c|}{Day of Month}
  & \multirow{2}{*}{Total} \\
 & \begin{tabular}[c]{@{}c@{}}Winter\\(Jan--Mar)\end{tabular}
 & \begin{tabular}[c]{@{}c@{}}Spring\\(Apr--Jun)\end{tabular}
 & \begin{tabular}[c]{@{}c@{}}Summer\\(Jul--Sep)\end{tabular}
 & \begin{tabular}[c]{@{}c@{}}Autumn\\(Oct--Dec)\end{tabular}
 & \begin{tabular}[c]{@{}c@{}}Night\\(00--06)\end{tabular}
 & \begin{tabular}[c]{@{}c@{}}Morning\\(06--12)\end{tabular}
 & \begin{tabular}[c]{@{}c@{}}Afternoon\\(12--18)\end{tabular}
 & \begin{tabular}[c]{@{}c@{}}Evening\\(18--24)\end{tabular}
 & \begin{tabular}[c]{@{}c@{}}Days\\1--10\end{tabular}
 & \begin{tabular}[c]{@{}c@{}}Days\\11--20\end{tabular}
 & \begin{tabular}[c]{@{}c@{}}Days\\21--31\end{tabular} & \\
\hline
\textbf{12h}   & 255 & 262 & 245 & 238 & 234 & 270 & 227 & 269 & 338 & 329 & 333 & \textbf{1,000} \\
\textbf{24h}   & 246 & 225 & 268 & 261 & 266 & 227 & 250 & 257 & 332 & 327 & 341 & \textbf{1,000} \\
\textbf{48h}   & 252 & 254 & 278 & 216 & 260 & 233 & 273 & 234 & 342 & 311 & 347 & \textbf{1,000} \\
\textbf{96h}   & 255 & 257 & 239 & 249 & 272 & 232 & 254 & 242 & 344 & 331 & 325 & \textbf{1,000} \\
\textbf{Total} & 1,008 & 998 & 1,030 & 964 & 1,032 & 962 & 1,004 & 1,002 & 1,356 & 1,298 & 1,346 & \textbf{4,000} \\
\bottomrule
\end{tabular}}
\medskip

\renewcommand\arraystretch{1.05}
\setlength{\tabcolsep}{8pt}
\captionof{table}{\textbf{Training hyperparameters and LUMI infrastructure.}
Best checkpoint: epoch 18, val RMSE\,=\,4.95, test RMSE\,=\,6.85\,\textmu g/m$^3$.}
\label{tab:s_hparams}
\resizebox{\textwidth}{!}{%
\begin{tabular}{@{} ll ll @{}}
\toprule
\textbf{Hyperparameter} & \textbf{Value} & \textbf{Hyperparameter} & \textbf{Value} \\
\midrule
Optimizer               & AdamW                      & Weight decay         & 0.05 \\
Learning rate           & $5\!\times\!10^{-5}$       & Min LR (cosine)      & $1\!\times\!10^{-6}$ \\
Warmup                  & 5 epochs                   & LR schedule          & cosine annealing \\
Max epochs              & 30                         & Best epoch           & 18 \\
Per-GPU batch           & 2                          & Effective batch      & 128 \\
Gradient clip           & 1.0 (global norm)          & Precision            & bfloat16-mixed \\
$\lambda_\text{pixel}$  & 1.0                        & $\lambda_\text{FFL}$ & 0.1 \\
$\lambda_\text{station}$ & 0.1                       & Drop-path            & 0.1 \\
Hotspot ratio           & 0.5                        & MLP ratio            & 4.0 \\
$E_0$ global            & 1000\,m                    & $E_0$ local          & 500\,m \\
Wind sectors            & 16                         & Patch size global    & $8\!\times\!8$ \\
Patch size local        & $16\!\times\!16$           & Tile overlap         & 64\,px \\
\midrule
\multicolumn{4}{@{}l}{\textit{LUMI supercomputer}} \\
Nodes                   & 4                          & GPUs                 & 64 AMD MI250X GCDs \\
Total HBM               & 4{,}096\,GB                & Interconnect         & Slingshot-11 \\
PyTorch                 & 2.2 (ROCm\,6.0.3)          & Strategy             & PyTorch Lightning DDP \\
Wall time               & $\approx$11\,h             & Total GPU-hours      & $\approx$860 \\
\bottomrule
\end{tabular}}
\medskip

\noindent The cosine annealing schedule with a minimum LR of $1\!\times\!10^{-6}$
provides gradual decay that prevents the model from converging prematurely on
the heterogeneous tile distribution. The hotspot ratio (0.5) controls the fraction
of training tiles sampled from high-pollution regions (Po Valley, Silesia, Rhine-Ruhr),
ensuring sufficient exposure to extreme concentration events. Drop-path
regularisation (0.1) is applied to both branches to prevent over-reliance on
individual tokens during co-training of the dual-branch architecture. The choice
of AdamW with weight decay (0.05) rather than standard Adam reflects the large
parameter space of the Swin Transformer blocks, where $\ell_2$ regularisation
is known to improve generalisation on spatially structured prediction tasks.

\renewcommand\arraystretch{0.85}
\setlength{\tabcolsep}{3pt}
\captionof{table}{\textbf{Input channel specification} (70 coarse + 5 local = 75 total).}
\label{tab:s_channels}
\begin{center}
\resizebox{0.72\textwidth}{!}{%
\begin{tabular}{@{} l l l l r @{}}
\toprule
\textbf{Variable} & \textbf{Symbol} & \textbf{Source} & \textbf{Norm.} & \textbf{\#ch} \\
\midrule
\multicolumn{5}{@{}l}{\textit{ERA5 surface, days $t$ and $t\!-\!1$}} \\
10-m wind              & $u_{10},v_{10}$ & ERA5 & z-score & 2 \\
2-m temperature        & $T_{2m}$        & ERA5 & z-score & 1 \\
Surface pressure       & $p_s$           & ERA5 & z-score & 1 \\
Precipitation          & $tp$            & ERA5 & z-score & 1 \\
\multicolumn{5}{@{}l}{\textit{ERA5 pressure levels (500,700,850,925,1000\,hPa), days $t$ and $t\!-\!1$}} \\
$u,v,T,z,q$ (×5 levels) & & ERA5 & z-score & 25 \\
\multicolumn{5}{@{}l}{\textit{CAMS analysis, days $t$ and $t\!-\!1$}} \\
PM$_{2.5}$, PM$_{10}$, NO$_2$, O$_3$, CO & & CAMS & z-score & 5 \\
\midrule
\multicolumn{4}{@{}l}{\textit{Total coarse: $(5+25+5)\times 2$}} & \textbf{70} \\
\midrule
\multicolumn{5}{@{}l}{\textit{Local tile (per $512\!\times\!512$)}} \\
GHAP PM$_{2.5}$ at $t$ and $t\!-\!1$ & & GHAP & $(x\!-\!15)/20$ & 2 \\
SRTM elevation         & $h$             & SRTM & raw\,m   & 1 \\
Latitude, Longitude    &                 & computed & degrees & 2 \\
\midrule
\multicolumn{4}{@{}l}{\textit{Total local}} & \textbf{5} \\
\bottomrule
\end{tabular}}
\end{center}
}

\begin{algorithm}[H]
\caption{\textbf{CRAN-PM forward pass (single tile).}
Global branch (lines 1--3) computed once per day and cached.
Local branch and cross-resolution attention (lines 4--8) run per tile.}
\label{alg:s_forward}
\begin{algorithmic}[1]
\Require $\mathbf{x}^g\!\in\!\mathbb{R}^{70\times168\times280}$,
  $\mathbf{x}^\ell\!\in\!\mathbb{R}^{5\times512\times512}$,
  $\tau\!\in\!\{1,2,3\}$,
  $x^{\ell,0}$ (today's PM$_{2.5}$ tile)
\Ensure $\hat{y} = x^{\ell,0} + \Delta$
\State Wind sectors $\{s_k\}$ from ERA5 $u_{10},v_{10}$ over $7\!\times\!7$ groups
\State $\mathbf{Z}^g \leftarrow \text{SwinBlocks}(\text{ElevAttn}(\text{UnfoldLinear}(\text{WindShuffle}(\mathbf{x}^g))))$
\State $\mathbf{Z}^g \leftarrow \text{WindUnShuffle}(\mathbf{Z}^g)$ \Comment{$735\!\times\!768$, cached once per day}
\State $\mathbf{Z}^\ell \leftarrow \text{SwinBlocks}(\text{ElevAttn}(\text{UnfoldLinear}(\mathbf{x}^\ell) + \mathbf{E}_\text{lead}(\tau)))$
\For{$i=1,2$} \Comment{cross-resolution bridge}
  \State $\mathbf{Z}^\ell \leftarrow \mathbf{Z}^\ell + \text{CrossAttn}(\mathbf{Q}\!=\!\mathbf{Z}^\ell,\;\mathbf{KV}\!=\!\text{Linear}(\mathbf{Z}^g);\;\mathbf{B}^\text{wind})$
  \State $\mathbf{Z}^\ell \leftarrow \mathbf{Z}^\ell + \text{FFN}(\mathbf{Z}^\ell)$
\EndFor
\State $\Delta \leftarrow \text{PixelShuffleDecoder}(\mathbf{Z}^\ell)$ \Comment{zero-init final conv}
\State \Return $\hat{y} = x^{\ell,0} + \Delta$
\end{algorithmic}
\end{algorithm}

\noindent The forward pass pseudocode highlights two design choices central to
CRAN-PM's efficiency. First, the Global Branch (lines 1--3) is computed once per
forecast day and its output $\mathbf{Z}^g$ is cached and reused across all 126
local tiles. This reduces the total number of global-branch forward passes from
$126\!\times\!N_\text{days}$ to $N_\text{days}$, saving approximately 99\% of
global-branch computation. Second, the cross-resolution attention (lines 5--7)
operates on compressed representations: local queries are $1{,}024\!\times\!512$
tokens and global key-values are $735\!\times\!512$ tokens, making the attention
matrix $1{,}024\!\times\!735 \approx 0.75$\,M entries per head, tractable on
modern GPUs without approximation. The wind bias $\mathbf{B}^\text{wind}$ is
computed from ERA5 sector assignments and is shared across all cross-attention
heads, adding negligible overhead while encoding physical advection direction
into every attention computation.

\end{document}